\documentclass[11pt]{article}

\usepackage[margin=1in]{geometry}
\usepackage[T1]{fontenc}
\usepackage{lmodern}
\usepackage{microtype}
\usepackage{amsmath,amssymb}
\usepackage{booktabs}
\usepackage{graphicx}
\usepackage{xcolor}
\usepackage{tikz}
\usepackage[hidelinks]{hyperref}
\usepackage{caption}
\usepackage{subcaption}
\usepackage{array}
\usepackage{url}
\usetikzlibrary{arrows.meta,positioning,shapes.geometric,calc,fit}

\graphicspath{{../figures/}}

\definecolor{kanblue}{HTML}{2F6BFF}
\definecolor{kanorange}{HTML}{E69F00}
\definecolor{kangreen}{HTML}{009E73}
\definecolor{kanpurple}{HTML}{7B61FF}
\definecolor{kanred}{HTML}{D55E00}
\definecolor{lightblue}{HTML}{EEF4FF}
\definecolor{lightorange}{HTML}{FFF3E0}
\definecolor{lightgreen}{HTML}{EAF7F2}
\definecolor{lightpurple}{HTML}{F1EEFF}
\definecolor{lightred}{HTML}{FFF0EA}

\newcommand{\modelname}[1]{\textsc{#1}}

\newcommand{\bpb}{\mathrm{BPB}}
\newcommand{\nls}{\mathrm{NLS}}
\newcommand{\grkan}{\modelname{GR-KAN}}
\newcommand{\mlpedge}{\modelname{MLPEdge}}

\newcommand{\kanlinear}{\modelname{KANLinear}}

\title{Kolmogorov--Arnold Networks for Small Language Models}
\author{%
  Felippe Alves \\
  TELUS Digital Research Hub -- CIAAM \\
  Institute of Mathematics, Statistics and Computer Science \\
  University of São Paulo \\
  \texttt{felippe.pereira@usp.br}\\
  \\
  Renato Vicente \\
  TELUS Digital Research Hub -- CIAAM \\
  Institute of Mathematics, Statistics and Computer Science \\
  University of São Paulo \\
  \texttt{rvicente@usp.br}
}
\date{\today}

\begin{document}
\maketitle

\begin{abstract}
Kolmogorov--Arnold Networks (KANs) replace fixed node activations with learned one-di\-men\-sion\-al edge functions, which makes them attractive both as interpretable components and as possible alternatives to transformer feed-forward networks. We evaluate these two premises separately. For interpretability, in a six-layer 10 million parameters scale B-spline KAN, all 884,736 feed-forward edge functions can be reconstructed and activity-ranked: most are active and nonlinear (87.8\% exceed the nonlinearity threshold \(\nls>0.1\), only 0.4\% are inactive), and the lowest-activity 20--25\% can be pruned for a negligible validation-loss increase, far outperforming random pruning---although a structured MLP neuron-pruning baseline tolerates comparable sparsity, so graceful FFN pruning is not unique to KANs. A grid-size sweep shows that the strongest functional-PCA and closed-form-fit summaries are properties of the low-capacity grid-2 basis (the per-edge function space is only six-dimensional there) rather than universal KAN behavior, and the full audit reproduces on a BabyLM grid-2 KAN, indicating the edge statistics are not specific to the one corpus. For the replacement premise, we move from a narrow custom corpus to a standardized testbed and find no consistent benchmark advantage. On BabyLM validation loss the gated and KAN-family feed-forward networks (SwiGLU, grouped Chebyshev, rational \grkan{}) all edge out a vanilla GELU MLP. Yet on standardized zero-shot benchmarks the architectures are benchmark-equivalent at the resolution of this study: across ten seeds and 59,875 BLiMP minimal pairs the accuracies span only 62.4--63.1\% with overlapping confidence intervals, EWoK sits at chance for every model, and the one robust effect---a \(+0.7\)-point \grkan{} edge on the main BLiMP suite---reverses on the BLiMP supplement, where the MLP is highest. Validation cross-entropy does not predict the benchmark ranking. Larger stress tests remain cautionary: a parameter-matched \mlpedge{} transformer underperforms an MLP on Wikitext-103, and corrected 286M-parameter \grkan{} runs remain below a SwiGLU MLP ClimbMix baseline even after a stabilized continuation. The strongest supported conclusion is therefore narrow but now standardized: KANs provide a practical, corpus-transferable audit interface for learned scalar transformations in small language models, while the tested KAN-family replacements show no consistent advantage over---and are benchmark-equivalent to---strong MLP baselines on standardized benchmarks, quality, and latency.
\end{abstract}

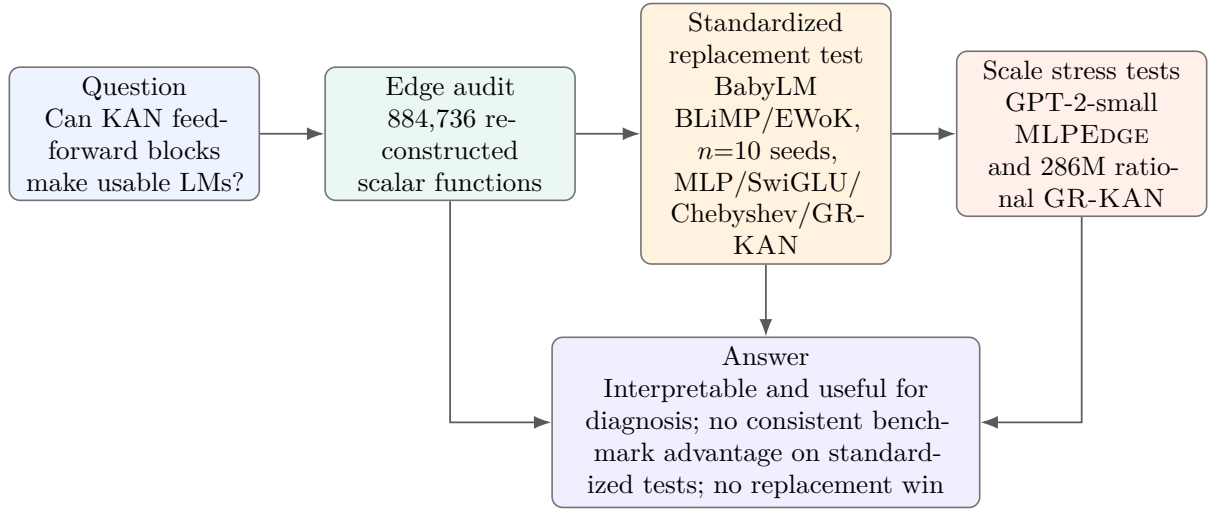
\begin{figure}[t]
\centering
\begin{tikzpicture}[
    node distance=0.95cm and 0.85cm,
    box/.style={rounded corners=4pt, draw=black!55, line width=0.6pt, align=center, minimum height=1.05cm, text width=3.05cm, font=\small},
    arrow/.style={-{Latex[length=2.3mm]}, line width=0.65pt, draw=black!65}
]
\node[box, fill=lightblue] (q) {Question\\Can KAN feed-forward blocks make usable LMs?};
\node[box, fill=lightgreen, right=of q] (interp) {Edge audit\\884,736 reconstructed scalar functions};
\node[box, fill=lightorange, right=of interp] (small) {Standardized replacement test\\BabyLM BLiMP/EWoK, \(n{=}10\) seeds,\\MLP/SwiGLU/\\Chebyshev/\grkan{}};
\node[box, fill=lightred, right=of small] (scale) {Scale stress tests\\GPT-2-small \mlpedge{} and 286M rational \grkan{}};
\node[box, fill=lightpurple, below=of small, text width=5.4cm] (answer) {Answer\\Interpretable and useful for diagnosis; no consistent benchmark advantage on standardized tests; no replacement win};
\draw[arrow] (q) -- (interp);
\draw[arrow] (interp) -- (small);
\draw[arrow] (small) -- (scale);
\draw[arrow] (interp.south) |- (answer.west);
\draw[arrow] (small.south) -- (answer.north);
\draw[arrow] (scale.south) |- (answer.east);
\end{tikzpicture}
\caption{Graphical abstract. The study separates the interpretability claim from the replacement claim. KAN edge functions can be reconstructed and summarized exhaustively in a small language model. On a standardized BabyLM testbed with ten seeds and the public evaluation pipeline, the tested KAN-family feed-forward networks show no consistent benchmark advantage over strong MLP baselines on BLiMP and EWoK, and larger stress tests remain cautionary.}
\label{fig:graphical-abstract}
\end{figure}

\section{Introduction}

Transformer language models rely on feed-forward sublayers that expand, transform, and compress each token representation after attention \cite{vaswani2017attention}. These blocks are usually implemented as dense affine maps with fixed nonlinearities, making them fast and well understood but difficult to inspect at the level of scalar computation. Kolmogorov--Arnold Networks offer a different design point: instead of applying a fixed activation at each hidden unit, a KAN learns univariate functions on edges between units \cite{liu2024kan}. This gives the architecture an explicit functional decomposition that could, in principle, make internal computations easier to reconstruct, prune, or symbolically summarize.

The central question is whether this interpretability premise remains useful when KANs are placed inside language models, and whether it comes with enough predictive or efficiency benefit to justify replacing standard MLP feed-forward blocks. The answer is not implied by prior KAN work. The original evidence for KANs is strongest on scientific regression and symbolic-discovery tasks \cite{liu2024kan,liu2024kan2}; KAN variant surveys identify many basis families and engineering gaps, but not a controlled language-model edge-function audit \cite{somvanshi2024survey,noorizadegan2025practitioners}. Kolmogorov--Arnold Transformer work shows that grouped rational KAN units can replace MLP blocks successfully in vision transformers \cite{yang2024kat}, but image classification differs from token prediction in activation statistics, sequence length, optimization objective, and throughput bottlenecks.

We present an empirical evaluation around three concrete questions. First, can learned KAN edge functions in a language model be reconstructed and analyzed exhaustively rather than sampled anecdotally? Second, when KAN-style feed-forward replacements are evaluated on a standardized small-language-model benchmark with adequate seeds, do they improve over strong MLP baselines on validation loss, standardized linguistic-competence benchmarks, or latency? Third, do promising small-scale variants survive stronger scale tests? The resulting picture is deliberately conservative. KANs are useful as interpretability instruments, because their scalar edge functions can be measured at the level of the full small model and reveal low-dimensional smooth structure. As general MLP replacements, the standardized evidence points to \emph{no consistent advantage}: on the BabyLM Strict-Small benchmark the tested KAN-family and gated-MLP feed-forward networks are benchmark-equivalent to a vanilla MLP at the resolution of this study---differences on BLiMP and EWoK are sub-1-point, ten-seed confidence intervals overlap, and the only robust main-BLiMP effect reverses on the supplement. An instructive negative accompanies this: validation cross-entropy does \emph{not} predict the benchmark ranking, so the small validation-loss advantages these variants show over the vanilla MLP do not translate into standardized-benchmark gains. The tested larger \mlpedge{} and corrected \grkan{} configurations likewise trail matched MLP controls or remain below available MLP endpoints after stabilization.

\section{Related Work}

\paragraph{KANs and efficient variants.}
The original KAN architecture uses learned B-spline edge activations and motivates a prune--symbolify workflow for scientific discovery \cite{liu2024kan}. KAN 2.0 extends this direction with multiplication nodes, tree conversion, and a compiler from symbolic formulas into KAN weights \cite{liu2024kan2}. These papers establish the interpretability motivation, but they do not measure what edge functions learn inside transformer language models. Later surveys and practitioner guides broaden the design space to FastKAN-style Gaussian/RBF bases, rational KANs, Fourier and wavelet bases, polynomial bases, and compact-support variants \cite{somvanshi2024survey,noorizadegan2025practitioners}. Efficient implementations such as FastKAN \cite{li2024fastkan} and EfficientKAN \cite{blealtan2024efficientkan} replace B-spline recursion with closed-form basis evaluation, closing part of the GPU-efficiency gap while preserving edge-wise interpretability. Wav-KAN \cite{bozorgasl2024wavkan} introduces wavelet bases that offer localized multi-resolution analysis for signal-like data. These variants identify a persistent GPU-efficiency gap for naive B-spline KANs: irregular basis evaluation and large edge-function tensors can make a parameter-matched KAN substantially slower than a dense MLP. This motivates our basis screen and our separation of parameter-matched from compute-aware claims.

\paragraph{Rational and grouped KANs.}
Kolmogorov--Arnold Transformer (KAT) replaces B-splines with group-shared rational activations evaluated by polynomial recurrences and reports strong ImageNet results at matched parameter/FLOP budgets \cite{yang2024kat}. This makes rational \grkan{} a natural candidate for language-model FFN replacement. However, KAT evaluates vision transformers rather than autoregressive token predictors. Language models put different pressure on memory bandwidth, activation outliers, sequence length, and benchmark sensitivity. Our corrected \grkan{} scale runs should therefore be read as a language-specific stress test of a promising vision-derived idea, not as a contradiction of the image-classification evidence. The broader rational-KAN literature, including rKAN's Pad\'{e} and rational-Jacobi bases \cite{aghaei2024rkan}, motivates treating denominator parameterization as an implementation-critical detail rather than an incidental coding choice.

\paragraph{Basis selection and small-LM evaluation.}
Gaussian/RBF KAN variants are attractive because they retain localized scalar functions without B-spline recursion, but scale selection is itself a hyperparameter: overly wide Gaussian bases collapse the first-layer feature matrix, while overly narrow bases undersmooth \cite{noorizadegan2026gaussian}. Chebyshev polynomial KANs provide a recurrence-friendly alternative with global support and simple GPU kernels \cite{sidharth2024chebyshev}. We include non-spline bases because a fair language-model study should not treat B-splines as the only KAN implementation. Small-language-model evidence is also sensitive to dataset, tokenizer, objective, and evaluation harness. DataComp-LM emphasizes controlled corpus and evaluation design for language-model comparisons \cite{li2024datacomp}; nanochat-style experiments similarly make data mixture, tokenizer lineage, and downstream CORE evaluation first-class experimental artifacts \cite{karpathy2025nanochat}. BabyLM and TinyStories benchmarks show that domain-specific small corpora can yield surprisingly strong transfer when evaluation is matched to the training distribution \cite{eldan2023tinystories,warstadt2023babylm}. TinyLlama \cite{zhang2024tinyllama}, MiniCPM \cite{hu2024minicpm}, and OLMo \cite{groeneveld2024olmo} establish strong open small-LM baselines with transparent training recipes, making them useful reference points for any architecture that claims to improve on small-scale dense transformers.

\paragraph{MLP baselines and interpretability.}
Modern transformer FFNs are not merely generic dense layers: GELU, SwiGLU, and related GLU variants are strong baselines that combine expressivity with efficient dense kernels \cite{shazeer2020glu}. SwiGLU in particular has become the default FFN in large production models because it outperforms GELU at matched parameter budgets \cite{chowdhery2022palm,touvron2023llama}. A replacement claim must therefore beat both quality and throughput, not only parameter count. Recent work on transformer FFN interpretability has shown that individual MLP neurons exhibit polysemanticity and superposition, making single-neuron interpretation difficult \cite{elhage2022superposition,gurnee2023finding}. Alternative approaches decompose FFN layers into interpretable basis directions or sparse feature dictionaries \cite{cunningham2023sparse,bricken2023monosemanticity}. KANs provide a different decomposition: instead of hidden-unit polysemanticity, they expose explicit scalar edge functions that can be directly visualized and curve-fitted. Interpretability work on additive models shows that learned scalar functions can be useful when their scope and assumptions are explicit \cite{agarwal2021nam}. KANs provide a richer edge-level decomposition than standard neural additive models, but high-fidelity curve fits should not be mistaken for symbolic laws. We therefore describe our closed-form fits as approximability over an observed activation domain, not as full symbolic regression; stronger claims would require a grammar search such as PySR and held-out domain tests \cite{cranmer2023pysr}.

\section{Methods}

\subsection{Study design and model families}

We evaluated KAN-style feed-forward layers across four regimes: (i) a controlled local comparison and interpretability source at GuppyLM scale, (ii) a standardized replacement benchmark on BabyLM Strict-Small, (iii) exhaustive edge-function reconstruction of the GuppyLM B-spline KAN, and (iv) larger scale stress tests. The first regime used GuppyLM, a small instruction-response language-model testbed built from \texttt{arman-bd/guppylm-60k-generic} \cite{guppylm60k}, a 60,000-example fish-personality corpus. The tokenizer is a project BPE tokenizer with vocabulary size 2,393, and the maximum sequence length is 128. The model has six transformer layers, hidden width 384, and six attention heads. Training uses assistant-token-only loss: prompt targets are set to \(-100\) and ignored by cross-entropy, which is essential for avoiding role-marker leakage and for explaining why validation cross-entropies are numerically low relative to ordinary all-token language modeling.

All principal GuppyLM rows use 8,000 optimizer steps, batch size 32, AdamW, cosine learning-rate decay, 200 warmup steps, gradient clipping, and seeds \(\{42,43,44\}\). The train split contains 57,000 examples and the test split contains 3,000 examples. After truncation and shifting, the train split contains 1,506,006 non-padding input positions and 860,389 assistant target tokens; the test split contains 79,231 input positions and 45,193 assistant target tokens. Simulating the actual 8,000-step dataloader gives 255,904 sampled examples per run and 3.862--3.864M assistant target tokens across seeds 42--44. These counts make clear that GuppyLM is a small interpretability testbed, not a broad open-web pretraining benchmark.

The GuppyLM model families were chosen to separate topology from basis choice. The MLP baseline uses a standard four-times expansion GELU feed-forward block. The B-spline KAN row uses explicit learned spline edge functions with grid size 2. The KAT-style row combines B-spline feed-forward layers with KAN-based attention projections and is therefore not an FFN-only isolation. \mlpedge{} preserves the additive edge-function topology but replaces each spline with a tiny scalar MLP. Corrected canonical \grkan{} uses group-shared rational activations with the Safe Pad\'{e} denominator. The square-denominator rational row tests a nearby denominator ablation. The Chebyshev row uses a grouped degree-3 polynomial basis with eight groups and tanh input normalization.

The second regime is a standardized replacement benchmark on BabyLM Strict-Small \cite{warstadt2023babylm}, a developmental corpus of roughly 10M words of child-directed and transcribed speech. Unlike GuppyLM, this regime uses standard all-token language modeling (no assistant masking) with a byte-level BPE tokenizer of vocabulary size 8,192 trained on the corpus and the same six-layer, width-384, six-head architecture (\(\approx\)13.8M parameters with a tied embedding). Each architecture is trained for 8,000 steps at batch size 32 with the GuppyLM optimizer settings, using ten seeds for the four critical rows (MLP, parameter-matched SwiGLU, grouped Chebyshev degree-3, and canonical rational \grkan{}), five seeds for supporting rows (square-denominator \grkan{}, B-spline KAN grid 2, \mlpedge{} \(h=8\)), and three for low-priority rows (KAT, \mlpedge{} \(h=5\))---61 runs in total. Trained checkpoints are wrapped as HuggingFace causal language models (a wrapper verified to reproduce the native model's logits bit-exactly) and evaluated with the public 2025 BabyLM evaluation pipeline \cite{babylm2025eval} in its zero-shot sentence-scoring mode on BLiMP \cite{warstadt2020blimp}, the BLiMP supplement, and EWoK \cite{ivanova2024ewok}. GLUE fine-tuning is reported as not run: the pipeline's sequence-classification path is incompatible with pure causal decoders (its default first-token pooling is degenerate for a causal model, and the last-token path raises a batch-indexing error), and obtaining GLUE numbers would require modifying the otherwise-unmodified official harness. The third regime reconstructs and analyzes learned B-spline KAN edge functions from the small GuppyLM KAN model. A six-layer, width-384 KAN FFN exposes \(384\times384=147{,}456\) scalar edge functions per layer and 884,736 across all FFN layers. The fourth regime stress-tests selected replacement ideas at larger scale: a parameter-matched \mlpedge{} transformer at GPT-2-small scale \cite{radford2019language} on Wikitext-103 \cite{merity2017pointer}, and corrected rational \grkan{} variants at approximately 286M parameters on ClimbMix using a nanochat-derived harness, including a stabilized \(g=4\) continuation to 5,040 steps.

\subsection{Metrics}

Language-model quality is measured by validation cross-entropy in GuppyLM and BabyLM, standardized zero-shot accuracy on the BabyLM benchmark suite (BLiMP, the BLiMP supplement, and EWoK), validation perplexity in the Wikitext-103 GPT-2-small experiment, and validation bits per byte in the 286M ClimbMix experiment. The BabyLM benchmarks are scored by the public pipeline as the fraction of items for which the model assigns higher total probability to the correct sentence (minimal pairs for BLiMP and the supplement; two-alternative completions for EWoK); chance is 50\% in every case. For BabyLM we report per-architecture means with 95\% confidence intervals over seeds and Welch two-sample \(t\)-tests against the MLP baseline, and note multiple-comparison sensitivity where relevant. The earlier fixed 16-prompt GuppyLM generation suite was only a heuristic behavior check; it is superseded here by the standardized BabyLM benchmarks and is not used as evidence. Latency is measured on same-device generation sweeps and interpreted only as local throughput evidence.

For compute reporting, we include validation quality, wall-clock, tokens/sec, precision, hardware, and MFU as separate quantities when logged. We do not collapse validation loss and time into a single scalar because the experiments are not compute-matched and backend utilization differs strongly across implementations. Peak VRAM and full FLOPs/token were not consistently logged for all experiment families and are listed as missing rather than inferred. Appendix~\ref{app:compute} provides the forward-pass FLOP estimation methodology.

Edge-function interpretability is measured by activity, nonlinearity score, roughness, functional principal component analysis, closed-form approximability, and threshold sensitivity. The nonlinearity score is
\[
\nls(f)=\frac{\lVert f-f_{\mathrm{affine}}\rVert_2}{\lVert f\rVert_2+\epsilon},
\]
where \(f_{\mathrm{affine}}\) is the best affine approximation to the reconstructed edge function over the sampled domain. Edges with \(\nls>0.1\) are counted as meaningfully nonlinear, and edges with activity at or below 0.01 are counted as effectively inactive.

\subsection{Edge-function reconstruction and analysis}

For each B-spline \kanlinear{} FFN layer, we reconstruct every edge function
\[
f_{j,i}(x)=W_{\mathrm{base}}[j,i]\,\mathrm{SiLU}(x)+\sum_k W_{\mathrm{spline}}[j,i,k]B_{i,k}(x),
\]
where \(B_{i,k}\) is the learned B-spline basis for input channel \(i\). The main 200-point audit evaluates each channel on its learned grid range, expanded by a 5\% margin, and computes metrics on the resulting curve. A repeat audit for seeds 42--44 used 120 points per curve to reduce runtime while preserving the same metric definitions. Functional PCA treats each layer as a curve matrix in \(\mathbb{R}^{(d_{\mathrm{out}}d_{\mathrm{in}})\times n}\), centers the curves on a common normalized domain, and applies SVD to extract the first four functional components. Closed-form approximability is assessed on the top-50 most active edges per layer by fitting the best function from the library \{linear, quadratic, cubic, sigmoid, tanh, \(x\cdot\mathrm{SiLU}(x)\)\} and reporting \(R^2\) on the observed domain. This is a smoothness and complexity check, not a claim of symbolic identification.

\subsection{A post-hoc univariate diagnostic for MLP FFNs}
\label{sec:mlp-edge}

To obtain a descriptive MLP reference, we reconstruct an ``effective edge function'' for each scalar input-output probe through the MLP FFN. A standard MLP FFN computes
\[
\mathrm{FFN}(x)=W_{\mathrm{out}}\,\phi(W_{\mathrm{in}}x+b_{\mathrm{in}})+b_{\mathrm{out}},
\]
where \(\phi\) is GELU. For each output channel \(j\) and input channel \(i\), we define the effective edge as the scalar function
\[
g_{j,i}(a)=W_{\mathrm{out}}[j,:]\,\phi\bigl(W_{\mathrm{in}}[:,i]\,a+b_{\mathrm{in}}\bigr)+b_{\mathrm{out}}[j],
\]
evaluated over the domain of observed activations \(a\) for input channel \(i\). This is a post-hoc diagnostic: it varies one input coordinate while holding the other pre-activation contributions at zero apart from the learned hidden bias. It is not a faithful circuit representation, because the MLP computes all channels simultaneously and the hidden-unit interactions are inseparable during forward propagation. We therefore use it only to ask whether this univariate probe of MLP input-output paths is nonlinear under the same NLS, fPCA, and closed-form tools used for KAN edges. The MLP NLS is computed against the best affine approximation to \(g_{j,i}\) on the same observed domain.

\section{Results}

\subsection{KAN edge functions are measurable, nonlinear, and low-dimensional}

The interpretability result is the clearest positive finding. In the B-spline KAN language model, all 884,736 feed-forward edge functions can be reconstructed and summarized. Most functions are not degenerate: in the 200-point audit, 87.8\% exceed the nonlinearity threshold and only 0.4\% are effectively inactive. The median nonlinearity score is 0.281 for reconstructed KAN edges. Under the post-hoc MLP diagnostic in Section~\ref{sec:mlp-edge}, analogous MLP input-output probes have median NLS 0.268, indicating that those probes are also mostly nonlinear on the observed activation domain. This comparison is descriptive rather than independent-sample causal evidence, because many probes share a layer, basis, and training context, and the MLP diagnostic is not an architectural edge decomposition. The near-equal medians also make the point that nonlinearity per se does not distinguish KAN edges from MLP paths; the relevant differentiator is \emph{faithfulness}---KAN edges are exact, separable scalar functions that can be pruned or edited (Section~\ref{sec:gridsweep} and the pruning result below), whereas the MLP effective edge is a post-hoc probe over a chosen activation domain with no such intervention.

\begin{figure}[tp]
\centering
\begin{subfigure}{0.86\linewidth}
\centering
\includegraphics[width=\linewidth]{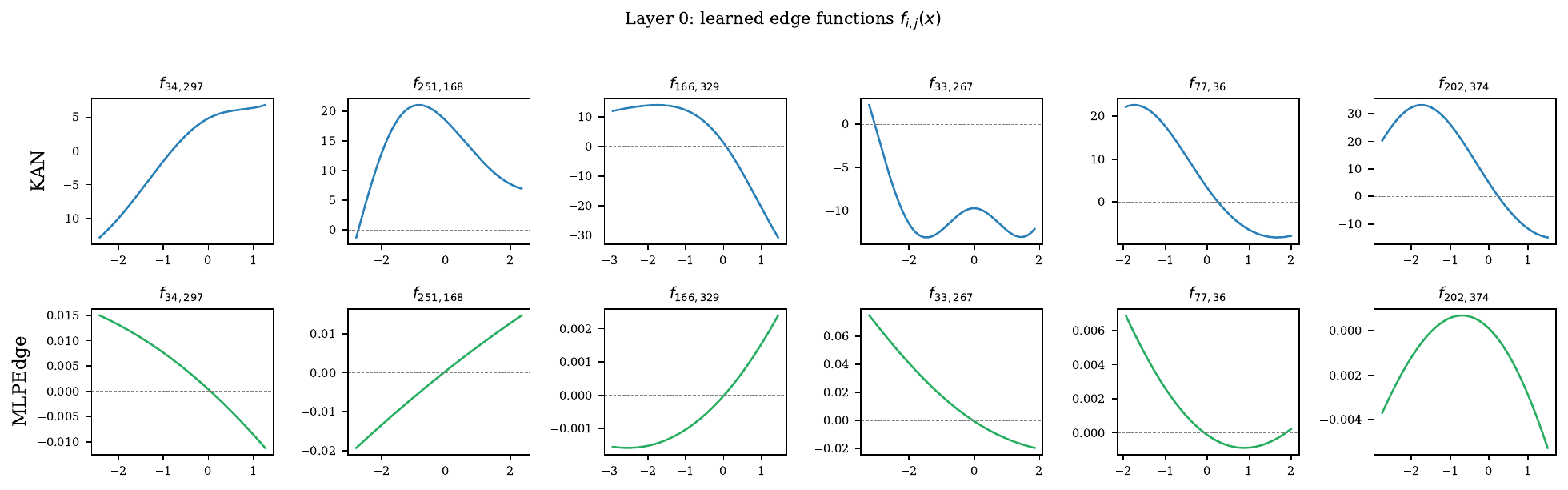}
\caption{Representative reconstructed edge curves.}
\end{subfigure}
\par\medskip
\begin{subfigure}{0.86\linewidth}
\centering
\includegraphics[width=\linewidth]{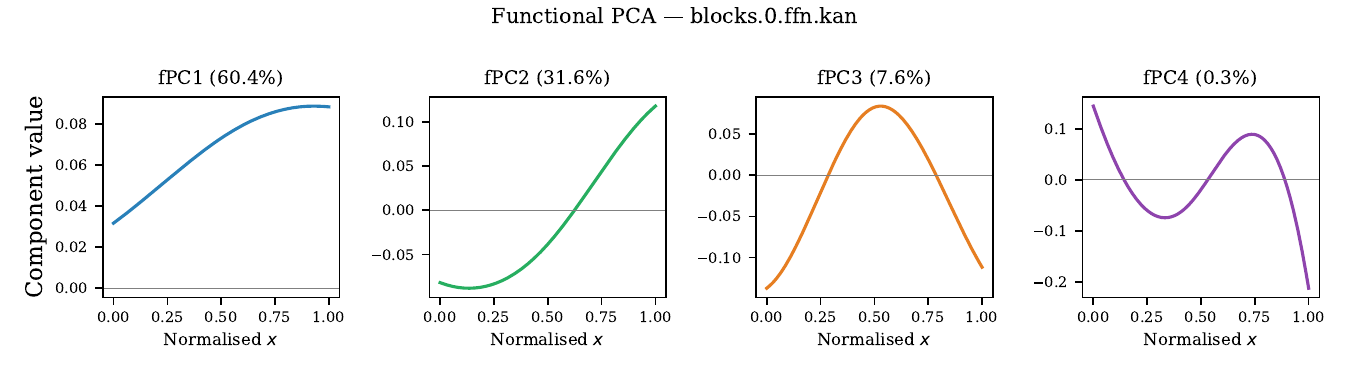}
\caption{Top functional principal components.}
\end{subfigure}
\par\medskip
\begin{subfigure}{0.86\linewidth}
\centering
\includegraphics[width=\linewidth]{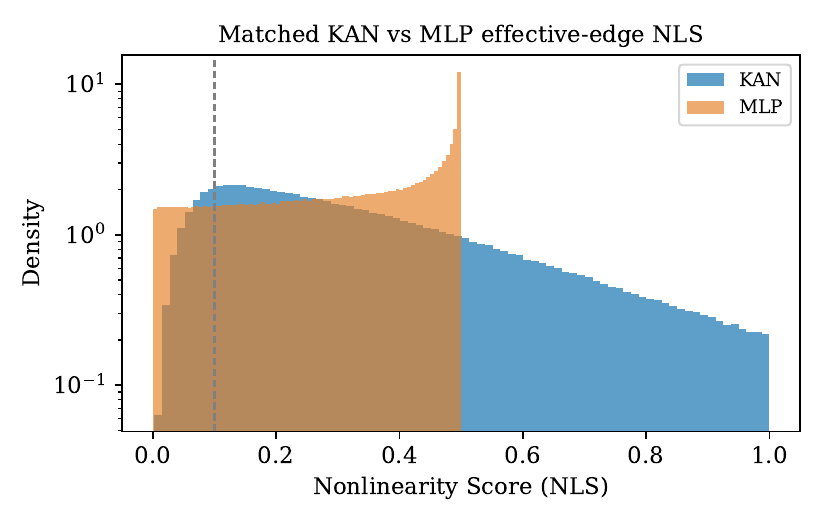}
\caption{KAN and MLP effective-edge NLS.}
\end{subfigure}
\caption{Edge-function evidence at GuppyLM scale. The audit reconstructs scalar functions rather than reporting only aggregate counts. Representative curves show smooth visualizable KAN edge functions; fPCA components show the dominant functional archetypes; and the NLS distribution shows that post-hoc MLP input-output probes are also predominantly nonlinear under the same univariate diagnostic.}
\label{fig:edge-evidence}
\end{figure}

Figure~\ref{fig:edge-evidence} replaces the aggregate-only bar chart with actual curve evidence. Representative edge functions are smooth and visually inspectable. The top fPCA components show the dominant archetypes: a smooth monotone component, an asymmetric smooth component, a bump-like component, and a small high-frequency perturbation. A functional PCA over edge curves shows that the top four components explain 99.9\% of curve variance in every layer. This figure is, however, largely a property of the grid-2 basis rather than of training: at grid size 2 each edge is a linear combination of only \(n_{\mathrm{basis}}+1=6\) shared functions (five B-splines plus the SiLU base), so the per-layer curve matrix has rank at most six and the top-four fPCA fraction is near-total for \emph{any} weights. A grid-size sweep (Section~\ref{sec:gridsweep}) confirms this: the top-four fraction falls from 99.9\% at grid 2 to 91\% at grid 5 and 75\% at grid 10, all at comparable validation loss. The low-dimensionality should therefore be read as a basis-capacity property of small-grid KANs, not as evidence that training uniquely compresses the function space.

Closed-form fitting tells a compatible story. Among the top 50 most active edges per layer, 93.3\% reach \(R^2>0.99\) against the six-function library on the observed domain. The dominant selected form is a cubic polynomial, with \(x\cdot\mathrm{SiLU}(x)\) and sigmoid accounting for the remainder. This does not identify symbolic laws, because smooth bounded curves are often well approximated by low-degree polynomials on a limited domain. It does show that the most active learned edge functions inhabit a low-complexity smooth regime where post-hoc summarization is plausible. As with the fPCA result, this high fit rate is strongest in the low-capacity regime: under the grid-size sweep (Section~\ref{sec:gridsweep}) top-50 closed-form coverage falls from 93--97\% at grid 2 to 18.7\% at grid 5 and 13.3\% at grid 10, all at comparable validation loss, so closed-form approximability is a property of the small grid-2 basis rather than a general feature of trained KAN language-model edges.

\paragraph{Control analysis: random initialization versus trained edges.}
We compared the trained audit against a randomly initialized (untrained) grid-2 KAN and against an initialization-variance sweep, and the two interpretability metrics behave differently under this control. The top-four fPCA fraction is \emph{not} diagnostic of training at grid 2: a freshly initialized grid-2 KAN (spline coefficients \(\sim\mathcal{N}(0,0.1)\), Kaiming-uniform base weight) already yields 99.9\% top-four variance, and sweeping the spline-weight initialization scale over two orders of magnitude leaves this figure unchanged, because the six-dimensional function space caps the achievable rank regardless of weights. (An earlier draft reported an 82.3\% random-init figure with high-frequency oscillations; that value is unreproducible at grid 2---five basis functions cannot oscillate---and is consistent only with a mismatched higher-grid configuration, whose random-init controls we measure at 91\% (grid 5) and 65\% (grid 10).) Closed-form approximability, by contrast, \emph{is} training-induced: the trained model reaches 93--97\% top-50 coverage against the six-function library, whereas matched random-init grid-2 edges reach only \(\approx\)63\%. We therefore retain only the supported claim---that training increases the smoothness (closed-form approximability) of the most active edges---and do not treat the low-dimensional fPCA structure as evidence of learning, since at grid 2 it is basis-imposed. The proper test of whether interpretability is learned rather than basis-imposed is the grid-size sweep below.

\paragraph{Grid-size sweep: interpretability versus basis capacity.}
\label{sec:gridsweep}
To separate learned structure from the grid-2 basis ceiling, we trained the same six-layer KAN at grid sizes 2, 5, 10, and 20 (\(n_{\mathrm{basis}}=5,8,13,23\)) and re-ran the full audit, extending the training horizon at the larger grids until validation loss converged. Both headline interpretability metrics degrade monotonically with basis capacity, and the degradation is present at matched validation loss. At grid 2 (validation loss 0.288) the top-four fPCA fraction is 99.9\% and top-50 closed-form coverage is 97\%; at grid 5 (validation loss 0.312) these fall to 91\% and 18.7\%; at grid 10 (validation loss 0.371, after extending training to 16{,}000 steps) they fall further to 75\% and 13.3\%. Because all three grid sizes reach comparable validation loss, this degradation reflects basis capacity rather than undertraining. Grid 20 reaches only 62\%/1\% but did not converge to comparable loss even at 16{,}000 steps (validation loss 3.02), which doubles as evidence that high-capacity B-spline KAN feed-forward layers are difficult to optimize at this scale. The exhaustive, smooth, low-dimensional, closed-form-fittable picture is therefore specific to the low-capacity (grid-2) regime: higher-capacity KAN feed-forward layers spend their additional basis functions on rougher, higher-dimensional edges that are not captured by four components or a small symbolic library. The interpretability claim should be scoped to small-basis KANs.

To address threshold and seed sensitivity, we repeated the aggregate edge audit on KAN grid-2 checkpoints trained with seeds 42, 43, and 44 using the same reconstruction code at 120 sample points. The nonlinear-edge rate at \(\nls>0.1\) ranges from 88.25\% to 88.43\%, inactive-edge rate ranges from 0.453\% to 0.455\%, and the top-four fPCA variance is above 99.92\% in every layer for every seed. Varying the nonlinearity threshold changes the absolute percentage, as expected, but not the conclusion that most edges are nonlinear: averaged across the three seeds, 97.34\% of edges exceed 0.05, 88.34\% exceed 0.10, 77.33\% exceed 0.15, and 66.80\% exceed 0.20. The audit is therefore robust at the level of aggregate edge statistics, although causal circuit claims would require separate interventions.

\paragraph{Cross-corpus replication on BabyLM.}
\label{sec:babylm-audit}
Because the full edge audit above is on GuppyLM---a narrow fish-personality corpus with a 2,393-token vocabulary and assistant-token masking---we repeated it on the BabyLM Strict-Small grid-2 B-spline KAN checkpoints (a standard developmental-text corpus, 8,192-token vocabulary, ordinary all-token language modeling; five seeds, identical reconstruction code and 200-point settings). The aggregate edge statistics reproduce almost exactly despite the very different token distribution and objective (Appendix Table~\ref{tab:babylm-audit}): pooled over five seeds the median NLS is \(0.294\) (GuppyLM \(0.281\)--\(0.293\)), \(88.4\%\) of edges are nonlinear (GuppyLM \(87.8\)--\(88.3\%\)), \(0.44\%\) are inactive (GuppyLM \(0.4\%\)), top-four fPCA variance is \(99.93\%\) (GuppyLM \(99.9\%\)), and top-50 closed-form coverage is \(92.8\%\) (GuppyLM \(93.3\%\)). The training-versus-basis split also reproduces: a random-initialized BabyLM grid-2 KAN reaches \(64.3\%\) closed-form coverage (versus \(\approx\)63\% for GuppyLM), confirming that closed-form approximability is training-induced while the near-total top-four fPCA variance is basis-imposed at grid 2 on both corpora. The interpretability audit is therefore not specific to the GuppyLM corpus; the grid-size scoping of Section~\ref{sec:gridsweep} remains the binding caveat, not the choice of training data.

\paragraph{The audit is actionable: activity-guided pruning.}
The edge audit is not only descriptive; the per-edge activity it measures identifies removable computation. Ranking all 884{,}736 grid-2 FFN edges by reconstructed-curve activity and zeroing the lowest-ranked fraction degrades validation loss gracefully---roughly 20--25\% of edges can be removed for a validation-loss increase below 0.005 nats, and 30\% for below 0.01---whereas zeroing the same fraction of \emph{randomly} chosen edges is far more damaging (validation loss 1.87 versus 0.355 at 50\% removed; even a 10\% random prune costs more than a 30\% activity-ranked prune).

\begin{figure}[t]
\centering
\includegraphics[width=0.72\linewidth]{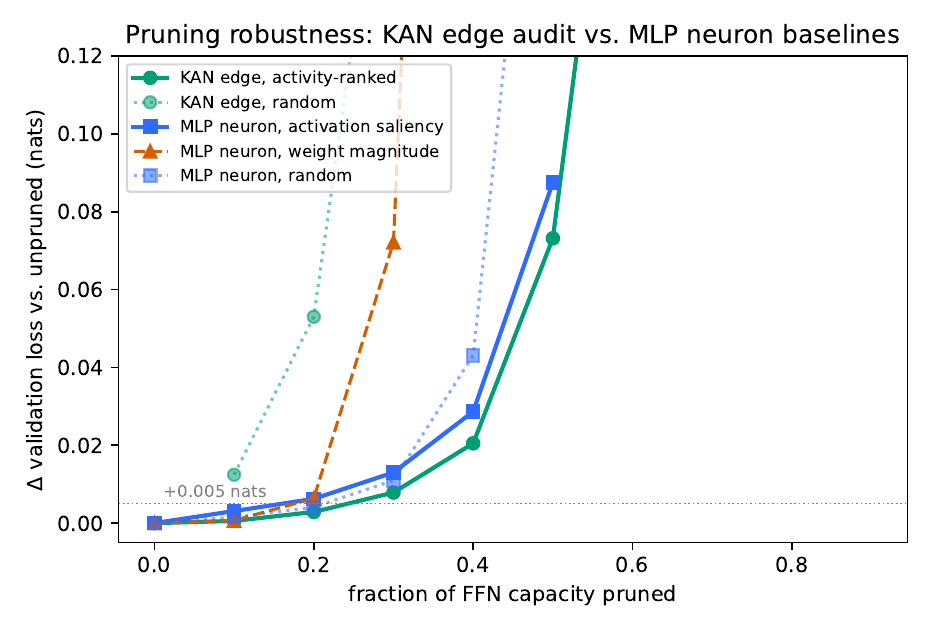}
\caption{Pruning robustness as a function of FFN-capacity sparsity, plotted as the increase in validation loss over the unpruned model so the KAN (grid-2 edges) and MLP (hidden neurons) curves are comparable despite different unpruned baselines. KAN activity-ranked pruning is the most graceful; a data-driven MLP activation-saliency baseline is comparable; MLP weight-magnitude pruning collapses past 20\%; random controls (dotted) are worst for both. Graceful pruning is therefore not unique to KANs, but the KAN activity metric---computed from reconstructed curves with no data---matches or beats the best MLP saliency at matched sparsity.}
\label{fig:prune-compare}
\end{figure}

To test whether this graceful pruning is specific to KANs or merely a generic property of redundant feed-forward layers, we ran a matched MLP neuron-pruning baseline on the GuppyLM MLP: the hidden neurons of the \(4\times\) GELU FFN were ranked by weight-magnitude saliency and by data-driven activation magnitude, and the lowest-ranked fraction was zeroed at the same FFN-capacity sparsities and evaluated on the same test split (Figure~\ref{fig:prune-compare}; Appendix Table~\ref{tab:prune-mlp}). A good MLP baseline---activation-magnitude pruning---also tolerates \(\approx\)20--25\% sparsity with small loss, so graceful FFN pruning is \emph{not} unique to KANs. At matched sparsity the KAN activity prune is comparable to, and modestly gentler than, the best MLP baseline in \(\Delta\) validation loss (KAN \(+0.008\) vs.\ MLP \(+0.011\) nats at 30\%; \(+0.021\) vs.\ \(+0.029\) at 40\%), and far more graceful than MLP weight-magnitude pruning, which collapses past 20\% (\(+0.07\) and \(+0.54\) nats at 30\% and 40\%). The audit's contribution is therefore not that pruning is uniquely possible for KANs, but that the per-edge activity metric---obtained zero-shot from the reconstructed curves, with no data forward passes---localizes removable computation at least as well as data-driven MLP saliency, while acting on faithful, separable scalar functions rather than entangled hidden units. The post-hoc MLP effective-edge probes of Section~\ref{sec:mlp-edge} admit no such structural prune; standard MLP neuron-pruning baselines, as just shown, remain a separate and competitive comparison.

\subsection{On a standardized benchmark the replacement variants show no consistent advantage over the MLP}
\label{sec:replacement}

\begin{figure}[t]
\centering
\includegraphics[width=0.98\linewidth]{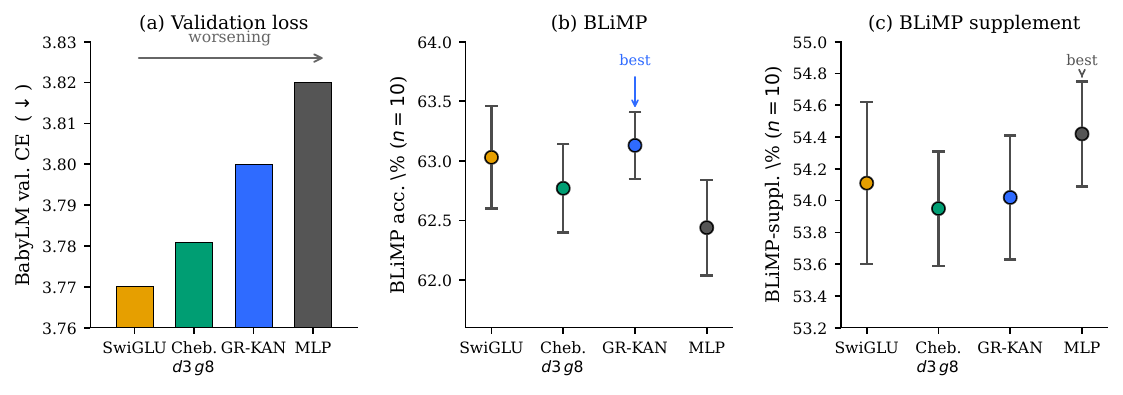}
\caption{Standardized BabyLM evaluation for the four critical architectures, ordered by validation loss (best to worst, left to right). (a) Validation cross-entropy worsens monotonically from SwiGLU to the MLP. (b) BLiMP accuracy (\(n=10\), 95\% confidence intervals) does not follow that order---the canonical \grkan{} is highest and all intervals overlap. (c) On the BLiMP supplement the ordering reverses and the MLP is highest. The dissociation between panel~(a) and panels~(b,c) is the central result: validation cross-entropy does not predict the standardized-benchmark ranking, and no architecture is consistently ahead.}
\label{fig:babylm}
\end{figure}

The clearest replacement result comes from the standardized BabyLM regime, where ten seeds per critical architecture and the public evaluation pipeline allow a properly powered comparison. Two findings stand out, and they pull in opposite directions. First, on BabyLM \emph{validation loss} the gated and KAN-family feed-forward networks all edge out the vanilla GELU MLP: mean validation cross-entropy is \(3.770\) for SwiGLU, \(3.781\) for Chebyshev, \(3.800\) for canonical \grkan{}, and \(3.820\) for the MLP, with seed-to-seed separations far larger than their spread (every pairwise difference is statistically reliable across the ten seeds, all \(p<10^{-10}\); exact values in Appendix~\ref{app:stats}). This reverses the GuppyLM ranking, in which the MLP led, and is direct evidence that validation-loss orderings among these architectures are domain-specific rather than intrinsic. Second, and decisively, this validation-loss advantage \emph{does not survive} on standardized linguistic-competence benchmarks. Table~\ref{tab:babylm} reports BLiMP and EWoK accuracy: across ten seeds and 59,875 BLiMP minimal pairs the four critical architectures span only \(62.4\)--\(63.1\%\), with overlapping confidence intervals, and EWoK sits at chance (\(\approx\)50\%) for every model. The architectures therefore show no consistent advantage on these benchmarks; they are benchmark-equivalent at the resolution of this study (Figure~\ref{fig:babylm}).

\begin{table}[t]
\centering
\caption{Standardized BabyLM Strict-Small zero-shot results, mean \(\pm\) 95\% confidence interval over ten seeds for the four critical architectures (all \(\approx\)13.8M parameters, tied embedding). BLiMP and the BLiMP supplement use the full evaluation set (59,875 and 5,218 minimal pairs); EWoK uses the fast set (1,100 two-alternative items; the full set requires a gated dataset). Validation cross-entropy is on the BabyLM held-out split. Chance is 50\% for all three benchmarks. Welch \(t\)-tests against the MLP are reported in the text. Rows ordered by validation loss.}
\label{tab:babylm}
\begin{tabular}{@{}lccccc@{}}
\toprule
Model & Params & Val CE & BLiMP & BLiMP-suppl. & EWoK \\
\midrule
SwiGLU-MLP & 13.83M & 3.770 & \(63.03\pm0.43\) & \(54.11\pm0.51\) & \(50.21\pm0.51\) \\
Chebyshev degree-3, \(g=8\) & 13.84M & 3.781 & \(62.77\pm0.37\) & \(53.95\pm0.36\) & \(49.73\pm0.47\) \\
Canonical rational \grkan{} & 13.84M & 3.800 & \(63.13\pm0.28\) & \(54.02\pm0.39\) & \(50.36\pm0.52\) \\
MLP-4x-GELU & 13.84M & 3.820 & \(62.44\pm0.40\) & \(54.42\pm0.33\) & \(49.77\pm0.55\) \\
\bottomrule
\end{tabular}
\end{table}

The one statistically robust effect is small and does not generalize. On the main BLiMP suite the canonical \grkan{} beats the MLP by \(+0.69\) points (\(t\)-test \(p=0.005\), the only comparison surviving a Bonferroni correction across the three non-MLP rows); SwiGLU is \(+0.58\) (\(p=0.037\)) and Chebyshev \(+0.32\) (\(p=0.19\), not significant). But on the BLiMP supplement the ordering reverses: the MLP is highest (\(54.42\)) and Chebyshev is significantly \emph{worse} than the MLP (\(-0.47\), \(p=0.04\)), with SwiGLU and \grkan{} also slightly below it. No architecture is consistently ahead across task families. The same picture holds when the supporting and low-priority rows are added on the fast benchmark set: BLiMP accuracy separates into a high-validation-loss cluster (B-spline KAN grid 2, \mlpedge{}, and KAT, all near \(59\%\)) and a low-validation-loss cluster (the four critical rows plus square-denominator \grkan{}, all near \(63\%\)), so BLiMP tracks validation loss only \emph{coarsely}---the \(\approx\)0.25-nat gap between clusters maps to a \(\approx\)4-point BLiMP gap---while within the top cluster the \(0.02\)--\(0.05\)-nat differences are within benchmark noise. The headline methodological consequence is that \textbf{validation cross-entropy does not predict the standardized-benchmark ranking}: SwiGLU has the best validation loss but is second on BLiMP, and Chebyshev has the second-best validation loss but is the weakest non-MLP row on BLiMP and significantly worse than the MLP on the supplement.

At the controlled GuppyLM scale that precedes this standardized test, the same architectures were screened on validation loss and latency, and that screen explains why these particular rows were carried forward. The controlled GuppyLM table shows that KAN usability depends strongly on basis and runtime. The GELU MLP baseline achieves a mean best validation loss of \(0.2770\pm0.0008\) across three seeds, and a parameter-matched SwiGLU MLP is stronger still at \(0.2763\pm0.0003\) with MLP-class latency (320.8\,ms). The strongest quality row among the tested KAN-family variants is the grouped Chebyshev degree-3 basis, which matches the GELU MLP mean at \(0.2770\pm0.0003\). This is a real small-scale quality result, but it is not a replacement win: the Chebyshev row neither beats the stronger SwiGLU baseline on quality nor approaches any dense baseline on latency, since its representative same-prompt latency (550\,ms) is substantially higher than either MLP (309--321\,ms). Corrected canonical \grkan{} and the square-denominator rational ablation are close to the MLP at \(0.2780\) mean validation loss, but neither improves the core objective. B-spline KAN, KAT, and \mlpedge{} rows trail the MLP on validation loss.

\begin{table}[t]
\centering
\caption{Controlled GuppyLM-scale replacement results. Validation loss is mean \(\pm\) sample standard deviation over three seeds. Training wall-clock is mean seconds for an 8,000-step run on Apple M4 Pro/MPS. Inference latency is mean generation latency (16 prompts, 64 tokens, seed-42) on the same local device. We report quality and time separately because these runs are not compute-matched and backend utilization differs across implementations. Both a GELU MLP and a parameter-matched SwiGLU MLP (hidden \(=8d/3\), three bias-free projections) are included as baselines. Parameter mismatches for B-spline (9.83M) and KAT (10.18M) are noted in the analysis text.}
\label{tab:guppylm}
\resizebox{\linewidth}{!}{%
\begin{tabular}{@{}lccccc@{}}
\toprule
Model & Parameters & Seeds & Best val loss & Train wall-clock (s) & Inf latency (ms) \\
\midrule
MLP-4x-GELU & 11.62M & 3 & \(0.2770\pm0.0008\) & 562.2 & 308.8 \\
SwiGLU-MLP & 11.60M & 3 & \(0.2763\pm0.0003\) & 600.3 & 320.8 \\
Chebyshev degree-3, \(g=8\) & 11.62M & 3 & \(0.2770\pm0.0003\) & 1245.7 & 550.2 \\
Canonical rational \grkan{} & 11.62M & 3 & \(0.2780\pm0.0005\) & 825.6 & 380.9 \\
Square-denom rational \grkan{} & 11.62M & 3 & \(0.2780\pm0.0009\) & 822.7 & 382.0 \\
B-spline KAN grid 2 & 9.83M & 3 & \(0.2883\pm0.0005\) & 1084.5 & 612.7 \\
\midrule
KAT grid 2 + KAT attn & 10.18M & 3 & \(0.2867\pm0.0006\) & 2785.4 & 887.7 \\
\midrule
\mlpedge{} \(h=5\) & 8.98M & 3 & \(0.2854\pm0.0005\) & 1861.8 & 470.9 \\
\mlpedge{} \(h=8\) & 11.64M & 3 & \(0.2862\pm0.0010\) & 763.8 & 535.8 \\
\bottomrule
\end{tabular}%
}
\end{table}

\paragraph{Profiler breakdown of the Chebyshev slowdown.}
Table~\ref{tab:profiler} decomposes forward-pass wall-clock time for the MLP and Chebyshev FFNs on Apple M4 Pro/MPS (batch 32, sequence 128). The isolated BasisKANFFN is \(9.2\times\) slower than the MLP FFN, yet the theoretical FLOPs difference is only \(\approx\)0.7\% because the two share identical dense projections (Appendix~\ref{app:compute}) and differ only in the activation. The gap is therefore kernel utilization, not arithmetic: the two matmuls take 2.0\,ms---the same as the MLP FFN---while the grouped basis evaluation accounts for roughly 90\% of the BasisKANFFN's 20\,ms, dominated by the per-group reshape, basis recurrence, and contraction rather than dense multiply--adds. Critically, compiling the \emph{entire} FFN block with \texttt{torch.compile} yields a \(2.0\times\) end-to-end speedup (to 10.1\,ms) but the compiled block is still \(4.6\times\) slower than the MLP FFN. Off-the-shelf fusion therefore substantially \emph{narrows} but does not \emph{close} the gap: at essentially matched FLOPs the basis activation remains memory- and launch-bound, so a competitive KAN FFN would require a hand-written fused basis kernel rather than compilation alone. The MLP FFN achieves \(71.8\%\) model MFU versus \(14.1\%\) for the full Chebyshev model. We confirmed the gap is not an MPS artifact by re-running the FFN microbenchmark on CUDA (RTX A6000, eager, batch 32, sequence 128): the MLP FFN runs in 0.60\,ms, the Chebyshev FFN in 3.32\,ms (\(5.5\times\) slower), and the rational \grkan{} FFN in 2.72\,ms (\(4.5\times\)). The relative gap is smaller than on MPS (\(9.2\times\)) because CUDA has better kernel coverage for the grouped basis evaluation, but the basis FFNs remain several times slower than the MLP at matched FLOPs even on a mature backend. (\texttt{torch.compile} was unavailable in the minimal benchmark image, so a fused-kernel CUDA figure remains future work; on MPS, whole-block compilation closed only about half the gap.)

\begin{table}[t]
\centering
\caption{Forward-pass profiler breakdown on Apple M4 Pro/MPS. Batch \(=32\), sequence \(=128\). Isolated FFN numbers are for a single FFN block; full-model numbers include attention, embeddings, and head. MFU uses a \(38\)~TFLOPS peak estimate for the M4 Pro GPU.}
\label{tab:profiler}
\begin{tabular}{@{}lccc@{}}
\toprule
Component & Median time (ms) & Tokens/s & MFU (\%) \\
\midrule
MLP full model & 25.6 & 160,000 & 71.8 \\
Chebyshev full model & 130.3 & 31,400 & 14.1 \\
\quad Slowdown (full model) & \multicolumn{3}{c}{\(5.1\times\)} \\
\midrule
MLP FFN (isolated) & 2.14 & 1,910,893 & 11.9 \\
BasisKANFFN (isolated) & 20.06 & 204,147 & 1.3 \\
\quad Slowdown (isolated FFN) & \multicolumn{3}{c}{\(9.4\times\)} \\
\midrule
\quad two dense projections (matmuls) & 2.03 & -- & -- \\
\quad grouped basis activation (both) & 18.4 & -- & -- \\
\midrule
BasisKANFFN + \texttt{torch.compile} (whole block) & 10.1 & -- & -- \\
\quad Compile speedup / residual vs MLP FFN & \multicolumn{3}{c}{\(2.0\times\) / still \(4.6\times\)} \\
\bottomrule
\end{tabular}
\end{table}

\paragraph{Statistical comparison (GuppyLM, local).}
The powered statistical comparison is the ten-seed BabyLM benchmark of Section~\ref{sec:replacement}; the GuppyLM screen below uses only three seeds and is reported as a local consistency check. With only three seeds, variance estimates are noisy, but Welch's two-sample \(t\)-tests between the MLP baseline and each KAN-family variant provide a formal check on whether observed differences are plausibly nonzero. The MLP versus Chebyshev comparison yields \(t(2.3)=0.00\), \(p=1.00\) (two-sided); the MLP versus canonical rational comparison yields \(t(2.9)=2.00\), \(p=0.12\); the MLP versus B-spline comparison yields \(t(2.9)=18.3\), \(p=0.0003\). Bootstrap 95\% confidence intervals for the mean difference (10,000 resamples) are \([-0.0015,+0.0015]\) for Chebyshev, \([-0.0024,+0.0004]\) for canonical rational, and \([+0.0107,+0.0113]\) for B-spline. The Chebyshev and canonical rational gaps are therefore not significantly different from zero at conventional thresholds given the low seed count, while the B-spline gap is clearly significant. A more definitive comparison would require five to ten seeds per row.

\paragraph{Benchmark dynamic range.}
The absolute validation losses are low (CE \(\approx\)0.28, perplexity \(\approx\)1.32), which raises the question of whether the task is too easy to separate architectures. It is not trivial: an add-\(0.1\) bigram baseline over the same assistant-token positions reaches CE 3.55 (perplexity 34.7) and a unigram 5.51 (perplexity 246), so the trained models capture roughly 3.3 nats of structure beyond a strong \(n\)-gram (a \(26\times\) perplexity reduction). However, all trained variants operate in the bottom \(\sim\)1\% of this range, where the inter-architecture spread (\(\approx\)0.011 nats) is small in absolute terms. The comparison is therefore meaningful---the models are far from a trivial floor---but the GuppyLM differences are small-margin separations near the achievable minimum, which reinforces that GuppyLM is an interpretability testbed rather than a high-dynamic-range replacement benchmark.

\paragraph{Parameter-count and runtime analysis.}
Parameter matching is necessary but not sufficient for fair architecture comparison. The B-spline KAN (9.83M) and KAT (10.18M) rows have fewer parameters than the 11.62M MLP, yet they trail the MLP on validation loss, so the observed gaps are not explained by parameter differences alone. The \mlpedge{} \(h=5\) row has only 8.98M parameters, but its higher loss (\(0.2854\)) relative to the parameter-matched \mlpedge{} \(h=8\) (\(0.2862\) at 11.64M) shows that the h=5 gap is partly architectural and partly parameter-driven. Runtime tells a separate story from parameter count. Chebyshev matches the MLP mean validation loss but takes 1245.7 seconds rather than 562.2 seconds for the same 8,000-step protocol and has higher same-device generation latency. Corrected canonical rational \grkan{} is closer to the MLP on runtime than Chebyshev, but still has slightly worse validation loss and higher generation latency. Because the runs are not compute-matched, these separate quality and time columns are more interpretable than a derived scalar.

\begin{figure}[t]
\centering
\includegraphics[width=0.88\linewidth]{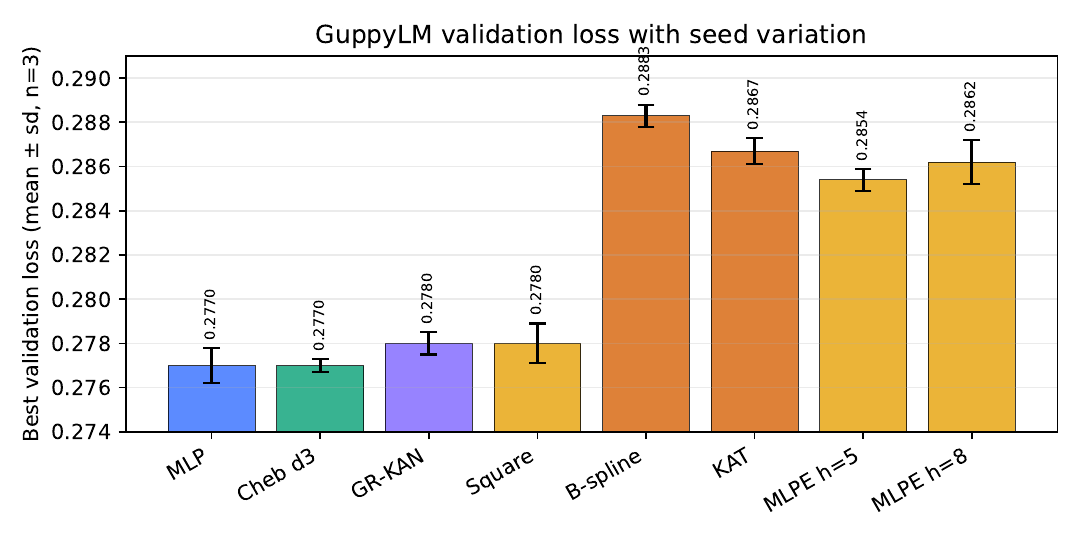}
\caption{Small-scale validation loss with seed variation. Error bars show sample standard deviation over three seeds. A grouped Chebyshev basis matches the MLP mean in this setting, rational \grkan{} rows are close, and B-spline/\mlpedge{} rows trail the MLP. Lower values are better.}
\label{fig:small-quality}
\end{figure}

Figure~\ref{fig:small-quality} visualizes the same quality comparison with error bars. Error bars show sample standard deviation over three seeds.

The KAT row should remain separated from FFN-only variants because it changes attention projections as well as the feed-forward layer. The earlier 16-prompt GuppyLM generation heuristic is superseded by the standardized BabyLM benchmarks of Section~\ref{sec:replacement} and is not used as evidence here; validation loss remains the primary GuppyLM metric.

\subsection{Scale stress tests remain cautionary for the tested replacement variants}

The larger experiments are single-run stress tests at limited training horizons, not definitive scaling laws. They are nevertheless useful because they test whether small-model observations survive more standard language-modeling conditions. At GPT-2-small scale, the tested \mlpedge{} \(h=8\) model is parameter-matched to a standard MLP transformer on Wikitext-103 but reaches validation perplexity 20.96, compared with 16.58 for the MLP control. It also has lower measured throughput, 48,921 tokens per second versus 54,820 tokens per second, and longer training wall-clock, 14.9 hours versus 13.3 hours on an RTX A6000. This result does not prove that all KAN variants fail at this scale, but it shows that the per-edge \mlpedge{} factorization does not transfer cleanly under this setup.

The corrected rational \grkan{} scale test is also cautionary, although the longer stabilized \(g=4\) run makes the result less one-sided than the original 2,520-step endpoint alone. At approximately 286M parameters and 2,520 training steps on ClimbMix, the MLP baseline reaches validation \(\bpb=0.8464\) and DCLM CORE 0.1540. The stronger corrected rational \grkan{} group setting, \(g=4\), reaches \(\bpb=0.8904\) and DCLM CORE 0.1284 at the same horizon. The \(g=8\) setting reaches \(\bpb=0.9030\) and DCLM CORE 0.1162. The group-count comparison favors \(g=4\) over \(g=8\), but both variants trail the MLP baseline on quality and throughput at 2,520 steps.

An unmatched stabilized \(g=4\) continuation improves \grkan{} BPB to \(0.8616\) by step 5,040, but because the MLP was not continued to the same horizon and the learning-rate schedule was reset, this is an optimization-stability observation rather than an architecture comparison; it does not alter the matched-horizon conclusion that both corrected \grkan{} group counts trail the MLP baseline at 2,520 steps (Appendix~\ref{app:continuation}). A credible scaling claim would require matched longer-horizon baselines, repeated seeds, and ideally an intermediate-scale sweep to test whether the gap closes with more steps or more data.

\begin{table}[t]
\centering
\caption{Larger-scale stress tests. The Wikitext-103 comparison is a GPT-2-small single-run experiment. The ClimbMix comparison uses corrected Safe Pad\'{e} \grkan{} runs at 286M parameters. The main matched-horizon comparison is at 2,520 steps; the stabilized \(g=4\) continuation reaches 5,040 steps but does not have a matched 5,040-step MLP continuation. Throughput, MFU, and wall-clock are not comparable across regimes because hardware, kernels, and datasets differ. Lower perplexity/BPB and higher CORE are better.}
\label{tab:scale}
\resizebox{\linewidth}{!}{%
\begin{tabular}{@{}llcccccc@{}}
\toprule
Regime & Model & Size & Primary metric & CORE / secondary & Throughput & Wall-clock & Precision \\
\midrule
Wikitext-103 & MLP transformer & 124M & PPL 16.58 & --- & 54,820 tok/s & 13.3h & CUDA default; not separately logged \\
Wikitext-103 & \mlpedge{} \(h=8\) & 125M & PPL 20.96 & --- & 48,921 tok/s & 14.9h & CUDA default; not separately logged \\
ClimbMix & SwiGLU MLP & \(\sim\)286M & BPB 0.8464 & CORE 0.1540 & \(\sim\)508K tok/s & run log metadata & BF16 \\
ClimbMix & Rational \grkan{} \(g=4\) & \(\sim\)286M & BPB 0.8904 & CORE 0.1284 & \(\sim\)290K tok/s & \(\sim\)1.23h metadata & BF16 \\
ClimbMix & Rational \grkan{} \(g=4\) stabilized continuation & \(\sim\)286M & BPB 0.8616 & --- & not separately logged & 149.2m & BF16 \\
ClimbMix & Rational \grkan{} \(g=8\) & \(\sim\)286M & BPB 0.9030 & CORE 0.1162 & \(\sim\)290K tok/s & \(\sim\)1.23h metadata & BF16 \\
\bottomrule
\end{tabular}%
}
\end{table}

The Safe Pad\'{e} denominator correction is important for interpreting these rows. Pre-correction rational kernels computed \(1+|b_0||x|+|b_1||x|^2+\cdots\), while the canonical formula is \(1+|b_0x+b_1x^2+\cdots|\). All pre-fix \grkan{} quality and scaling claims are excluded from this manuscript. The corrected formula is necessary for honest evaluation, but it is not sufficient to make the tested \grkan{} variants competitive with the MLP baseline.

\section{Discussion}

The strongest defensible contribution is practical edge-function auditing. The B-spline KAN model exposes a full feed-forward sublayer as scalar functions, and those functions can be reconstructed, summarized, compared, and visualized at the level of the full small model. This gives a concrete form to the KAN interpretability promise in a language model: the learned functions are mostly active, mostly nonlinear, strongly low-dimensional under fPCA, and smooth enough for high-fidelity approximation by a small function library on the observed domain. These properties characterize the grid-2 model audited here; the grid-size sweep (Section~\ref{sec:gridsweep}) shows that the low-dimensional fPCA structure and closed-form approximability weaken as basis capacity grows, so they should be read as properties of small-basis KANs rather than of KAN feed-forward layers in general. With that scope, the audit is also \emph{actionable}: the per-edge activity it measures localizes removable computation, and activity-ranked pruning removes a fifth to a quarter of the edge functions with negligible loss. These properties make small-basis KANs useful as diagnostic models, controlled interpretability sandboxes, and architecture probes for understanding what scalar transformations language-model feed-forward layers learn.

The replacement story is more constrained, and the standardized evidence sharpens it from ``promising but unproven'' to ``no consistent advantage.'' On the BabyLM benchmark the tested KAN-family and gated feed-forward networks are benchmark-equivalent to the MLP at the resolution of this study: BLiMP accuracies overlap within ten-seed confidence intervals, EWoK is at chance for every model, and the only robust effect---a sub-one-point canonical \grkan{} edge on the main BLiMP suite---reverses on the BLiMP supplement, where the MLP is highest. The most instructive negative is that validation cross-entropy does not predict this ranking. The same variants that edge out the MLP on BabyLM validation loss---and the SwiGLU baseline that leads on validation loss---do not lead on the standardized benchmarks, and the GuppyLM and BabyLM validation-loss orderings disagree with each other. This both undercuts validation loss as a proxy for architecture quality at this scale and explains why a controlled validation-loss screen, however careful, cannot by itself support a replacement claim. Even benchmark-equivalent quality is in any case not sufficient when the MLP remains faster and simpler. A \texttt{torch.profiler} breakdown of the Chebyshev forward pass on MPS (Table~\ref{tab:profiler}) shows that the grouped basis evaluation---which is not fused into optimized kernels---consumes roughly 90\% of FFN wall-clock time at FLOPs essentially matched to the MLP. Compiling the whole FFN block with \texttt{torch.compile} halves its latency but leaves it \(4.6\times\) slower than the MLP FFN, so the gap is a kernel-fusion problem that off-the-shelf compilation narrows but does not close---closing it would require a dedicated fused basis kernel. The larger stress tests point the same way. \mlpedge{} is helpful for isolating the value of edge-wise scalar functions without spline machinery, but its GPT-2-small result shows that small-scale competitiveness within the B-spline/\mlpedge{} subset can disappear under a more standard language-modeling stress test. Corrected rational \grkan{} appears less damaging than \mlpedge{} at 286M parameters, but at the matched 2,520-step horizon both group counts still trail the MLP baseline; an unmatched stabilized continuation reduces its BPB further without changing that matched-horizon conclusion (Appendix~\ref{app:continuation}).

\paragraph{Required evidence for scaling claims.}
A credible scaling claim would require: (a) matched training horizons of at least 10,000 steps at 286M parameters, (b) repeated seeds to estimate variance, and (c) an intermediate-scale sweep (e.g., 50M--150M parameters) to test whether the gap closes with more data or more capacity. The stabilized \(g=4\) continuation addresses part of the horizon criticism by extending corrected \grkan{} evidence to 5,040 steps and reducing validation BPB from 0.8904 to 0.8616. It does not satisfy the scaling bar because there is no matched 5,040-step MLP continuation, the learning-rate schedule was reset, and the result is still a single run. We therefore treat the ClimbMix evidence as a cautionary stress test rather than evidence for or against an asymptotic scaling law.

\begin{figure}[t]
\centering
\includegraphics[width=0.95\linewidth]{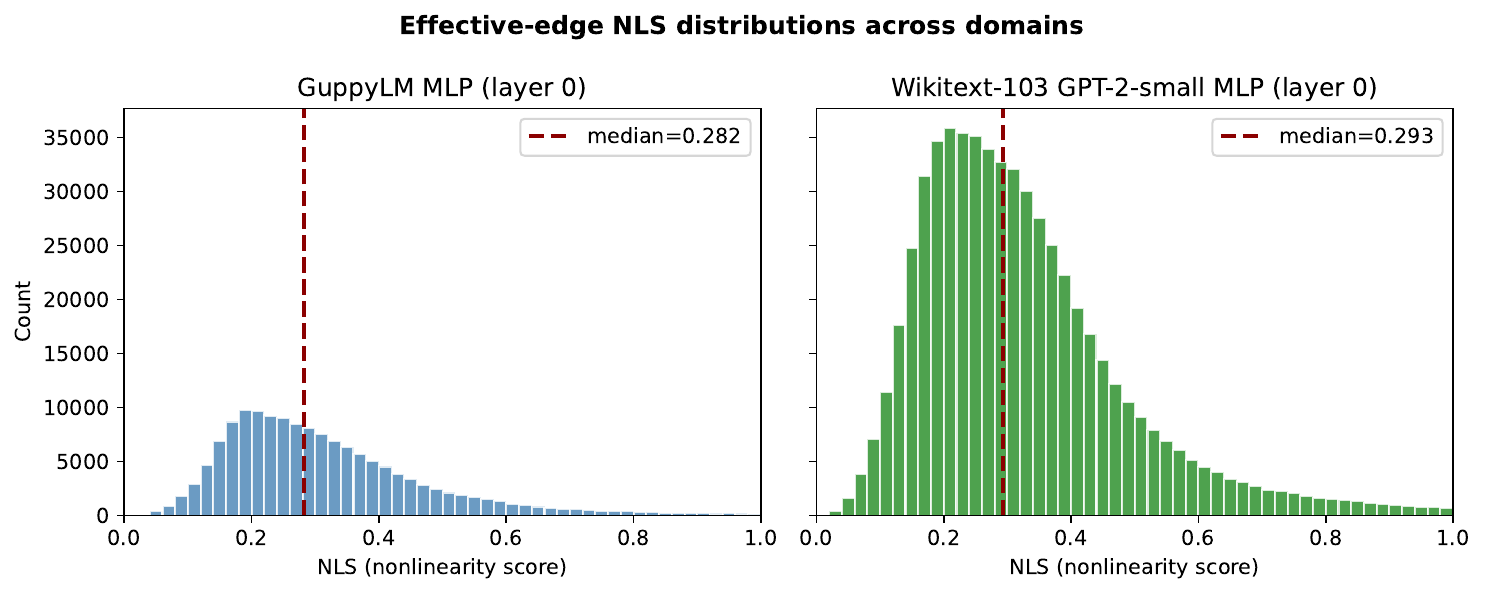}
\caption{Post-hoc MLP effective-edge NLS distributions for the first FFN layer of the GuppyLM MLP (left, 384\(\times\)768 probes) and the Wikitext-103 GPT-2-small MLP (right, 768\(\times\)3072 probes). Dashed vertical lines show medians (\(0.282\) and \(0.293\) respectively). The similarity indicates that these univariate MLP probes are predominantly nonlinear across corpora; it does not establish that MLPs contain architectural scalar edge functions comparable to KAN edges.}
\label{fig:nls-cross-domain}
\end{figure}

\paragraph{Domain limitations and the division of testbeds.}
The two premises are anchored primarily on different corpora, by design. The \emph{replacement} claim rests on BabyLM Strict-Small, a standard developmental-text benchmark with an 8,192-token vocabulary, ordinary all-token language modeling, and the public evaluation suite---a substantial step up in breadth and standardization from the earlier GuppyLM screen. The full \emph{interpretability} audit was developed on the GuppyLM grid-2 B-spline KAN---a narrow fish-personality instruction-response corpus with only 2,393 vocabulary tokens and assistant-token-only masking---but its aggregate edge statistics replicate closely on the BabyLM grid-2 KAN (Section~\ref{sec:babylm-audit}; Appendix Table~\ref{tab:babylm-audit}), so the audit's quantitative properties are not artifacts of the GuppyLM corpus. The methodology---exhaustive reconstruction, NLS scoring, fPCA, closed-form fitting---is in any case domain-agnostic, and the binding scope limitation is basis capacity (Section~\ref{sec:gridsweep}), not the choice of training data. As one adjacent cross-corpus check on the MLP side, we repeated the post-hoc MLP NLS diagnostic on a Wikitext-103-trained GPT-2-small checkpoint: the median NLS is \(0.293\), compared with \(0.282\) for the GuppyLM MLP (layer 0 in both cases), indicating that these univariate MLP probes are predominantly nonlinear across corpora (Figure~\ref{fig:nls-cross-domain}). What remains untested is the audit on open-web pretraining at larger scale and at higher grid sizes, where Section~\ref{sec:gridsweep} predicts the low-dimensional summaries would weaken; domain-specific small corpora like TinyStories produce surprisingly strong transfer when evaluation is matched \cite{eldan2023tinystories}, but neither corpus audited here is open-web.

\paragraph{Comparison against a strong SwiGLU baseline.}
Because SwiGLU typically outperforms GELU at matched parameter budgets \cite{chowdhery2022palm}, we added a parameter-matched small-scale SwiGLU MLP (hidden \(=8d/3\), three bias-free projections; 11.60M parameters) so the replacement comparison is made against a modern FFN, not only the older GELU block. The SwiGLU baseline is the strongest quality row in Table~\ref{tab:guppylm} at \(0.2763\pm0.0003\), slightly better than the GELU MLP and the Chebyshev tie, while retaining MLP-class latency (320.8\,ms). Consequently no tested KAN-family variant beats the SwiGLU baseline on either quality or latency in the GuppyLM screen: the small-scale Chebyshev ``match'' is a match only to the weaker GELU baseline, and the strongest dense baseline remains both better and faster than every KAN variant tested. The standardized BabyLM result completes this picture from the other direction. There SwiGLU again has the best validation loss, yet its BLiMP and EWoK accuracy is within the sub-1-point, overlapping-CI spread of both the GELU MLP and the KAN-family rows. Even the strongest dense feed-forward network therefore does not convert a validation-loss lead into a standardized-benchmark lead---the cleanest single illustration that these architectures are benchmark-equivalent at this scale and that the relevant differences are interpretability and engineering, not linguistic-competence accuracy.

These results clarify how future KAN-LLM work should be evaluated. Small custom datasets are useful for interpretability because they keep models small enough for exhaustive edge analysis, but they are not sufficient for architecture claims, which need matched tokenization, matched objectives, multiple seeds, direct comparison to a strong SwiGLU or Gated MLP baseline, standardized downstream benchmarks, and compute reporting that distinguishes parameter matching from compute matching. The BabyLM regime here meets most of that bar at small scale---a standard objective, ten seeds, a parameter-matched SwiGLU baseline, and a public benchmark suite---and under those conditions the answer is no consistent advantage. What remains open is the same standardized comparison at larger scale and with matched longer-horizon baselines. Evaluation should also separate KAN basis design from unrelated attention changes: the KAT-style GuppyLM row is informative, but because it modifies attention as well as the feed-forward layer, it cannot alone support an FFN-only speed or quality claim.

The main limitation of this study is that the standardized evidence is well powered at small scale but still sparse at larger scale. The BabyLM regime supplies ten seeds per critical architecture on a public benchmark, so the small-scale no-advantage conclusion is statistically grounded; the larger experiments are not. The GPT-2-small \mlpedge{} result is a single negative case study, and the 286M rational \grkan{} evidence covers two completed group counts at 2,520 steps plus one stabilized \(g=4\) continuation to 5,040 steps, with no matched longer-horizon MLP baseline. EWoK is at chance for every model, so it constrains world-knowledge comparison only weakly at this size, and GLUE could not be run on the unmodified pipeline. Consequently, the paper should not be read as proving that KANs cannot become competitive language-model components. It shows instead that under a standardized small-scale evaluation the tested KAN-family FFNs show no consistent advantage over strong MLP baselines, and that demonstrating a replacement advantage would require stronger quality, speed, and scaling results than are presently available.

\section{Conclusion}

KANs can be used productively in language-model research when the goal is to expose and analyze learned scalar transformations. In the audited low-capacity grid-2 models---demonstrated on GuppyLM and, by replication, on BabyLM---the edge functions are reconstructable, mostly nonlinear, and compressible into smooth functional archetypes, though the grid-size sweep shows the compressibility is a property of the small basis rather than of KAN feed-forward layers in general. This validates the interpretability premise at the level of the full small model. The architecture-replacement premise remains unproven, and on a standardized benchmark it resolves to no consistent advantage. On BabyLM Strict-Small with ten seeds and the public evaluation pipeline, the tested KAN-family and gated feed-forward networks are benchmark-equivalent to the MLP on BLiMP and EWoK; the one robust effect---a sub-one-point \grkan{} edge on the main BLiMP suite---reverses on the BLiMP supplement, where the MLP is highest; and validation cross-entropy does not predict the benchmark ranking, so the small validation-loss advantages these variants show do not constitute a replacement win. Across larger stress tests, the evaluated \mlpedge{} and corrected rational \grkan{} configurations either underperform matched MLP controls or, in the stabilized \grkan{} continuation, narrow the gap without producing a matched-horizon win. The practical recommendation is therefore to treat KANs as interpretable experimental components and basis-design probes, not yet as drop-in replacements for modern transformer FFNs.

\section*{Data and artifact availability}

Code, scripts, per-seed BabyLM evaluation reports, and small-scale checkpoints will be released in an anonymized archive before review (URL provided through the submission system) and permanently archived on Zenodo with a citable DOI for the camera-ready version. The archive is a single self-contained repository containing frozen vendor copies of the code used for the GuppyLM training/audit/pruning experiments, the BabyLM export and evaluation harness, and the nanochat/\grkan{} scale stress tests, together with a provenance manifest (\texttt{manifest.json}) that records the exact repository revisions, commands, seeds, tokenizer artifacts, and the evaluation report backing every figure and table. The corrected rational-kernel revisions and the Safe Pad\'{e} denominator correction are documented in Appendix~\ref{app:coderev}, and the full command, seed, environment, and checkpoint provenance is in Appendix~\ref{app:reprod}. The BabyLM regime additionally includes a sequential training launcher for the 61-run seed matrix, a \texttt{trust\_remote\_code} HuggingFace wrapper that reproduces each native model's logits bit-exactly, and evaluation scripts that drive the public 2025 evaluation pipeline \cite{babylm2025eval} in its zero-shot mode. The paper intentionally distinguishes released scientific artifacts from infrastructure-specific pod identifiers. Dataset licenses are reported in Appendix~\ref{app:reprod}. The GuppyLM corpus is MIT licensed; Wikitext-103 is distributed under CC BY-SA 3.0/GFDL terms; the ClimbMix source lineage is traced to NVIDIA Nemotron-ClimbMix under CC BY-NC 4.0 terms for research and development use.

\appendix

\section{Reproducibility and implementation details}
\label{app:reprod}

\subsection{Code revisions and denominator correction}
\label{app:coderev}

The GuppyLM and Wikitext-103 runs use the \texttt{kan-guppylm} codebase at commit \texttt{3c8ea92}; the GuppyLM MLP pruning baseline (Section~\ref{sec:replacement}) and the BabyLM grid-2 edge audit (Section~\ref{sec:babylm-audit}) use the same codebase. The ClimbMix \grkan{} runs use the \texttt{nanokan} harness wrapping nanochat, with nanochat commit \texttt{dc54a1a}; the stabilized 5,040-step continuation records \texttt{nanokan} reference \texttt{8b21df237143} and rational-kernel reference \texttt{41a20b5}. The corrected rational activation is pinned to a Safe Pad\'{e} implementation whose canonical denominator is
\[
Q(x)=1+|b_0x+b_1x^2+b_2x^3+b_3x^4|.
\]
The retracted pre-fix formula was \(1+|b_0||x|+|b_1||x|^2+\cdots\). Pre-fix \grkan{} checkpoints are not used as evidence. Gradient consistency and smoke tests were run before the corrected d12 jobs; the resulting corrected runs are the only \grkan{} scale rows reported in the main text.

\subsection{Environment and hardware}

\begin{table}[htbp]
\centering
\caption{Recorded run environments.}
\label{tab:env}
\resizebox{\linewidth}{!}{%
\begin{tabular}{@{}lccc@{}}
\toprule
Component & GuppyLM Track A & \mlpedge{} Wikitext-103 & \grkan{} ClimbMix d12 \\
\midrule
OS & Darwin 25.4.0 arm64 & Ubuntu 22.04 & Ubuntu 22.04 \\
Python & 3.13.5 & 3.10.12 & 3.10.12 \\
PyTorch & 2.12.0 (MPS) & 2.4.1 (CUDA 12.4) & 2.5.1 (CUDA 12.4) \\
Accelerator & Apple M4 Pro, MPS & RTX A6000 48GB & H100 80GB train; RTX PRO 4500 eval \\
Precision & float32/MPS default & CUDA default not separately logged & BF16 mixed precision, Flash Attention 3 \\
Primary compute fields & wall-clock, MPS latency & train time, tok/s & BPB, CORE, tok/s, metadata, BF16 logs \\
\bottomrule
\end{tabular}%
}
\end{table}

Peak VRAM and architecture-specific FLOPs/token were not logged consistently for all runs. We therefore report throughput and wall-clock as observed compute proxies and avoid claiming compute-matched parity. Appendix~\ref{app:compute} provides estimated FLOPs/token and training-efficiency proxies where computation could be reconstructed from the architecture definitions.

\subsection{Datasets, tokenizers, and token budgets}

\begin{table}[htbp]
\centering
\caption{Dataset and tokenizer provenance.}
\label{tab:data}
\resizebox{\linewidth}{!}{%
\begin{tabular}{@{}lll@{}}
\toprule
Regime & Dataset and license & Tokenizer / objective \\
\midrule
GuppyLM & \texttt{arman-bd/guppylm-60k-generic}, MIT, DOI \texttt{10.57967/hf/8339} & BPE vocab 2,393; assistant-token-only loss \\
BabyLM Strict-Small & BabyLM developmental corpus (\(\sim\)10M words) \cite{warstadt2023babylm} & byte-level BPE vocab 8,192; standard all-token LM objective \\
Wikitext-103 & \texttt{Salesforce/wikitext}, \texttt{wikitext-103-v1}, CC BY-SA 3.0/GFDL & GPT-2 BPE vocab 50,257; standard LM objective \\
ClimbMix & nanochat ClimbMix lineage to NVIDIA Nemotron-ClimbMix, CC BY-NC 4.0 & RustBPE vocab 32,768; training tokenizer reused for eval \\
\bottomrule
\end{tabular}%
}
\end{table}

For GuppyLM, the train split contains 57,000 examples, 1,506,006 shifted input positions, and 860,389 assistant target tokens after masking. The test split contains 3,000 examples, 79,231 input positions, and 45,193 assistant target tokens. For 8,000-step runs with batch size 32, the observed sampled-example count is 255,904 because the final dataloader batch is incomplete before cycling. Assistant target-token counts are 3,863,378 for seed 42, 3,862,385 for seed 43, and 3,863,730 for seed 44. The Wikitext-103 scale run uses 20,000 steps with effective batch size 128 and context length 1,024. The original ClimbMix d12 comparison uses 2,520 steps, sequence length 2,048, and total batch 524,288 tokens per step, for approximately 1.32B training tokens. The stabilized \(g=4\) continuation resumes from a stabilized step-3,000 pilot and runs to step 5,040, corresponding to a nominal 2.64B-token horizon at the same batch-token setting; because the learning-rate schedule is reset, it is reported as a continuation stress test rather than a seamless extension of the original 2,520-step run.

\subsection{Exact GuppyLM command patterns}

All GuppyLM principal rows use 8,000 steps, batch size 32, and seeds 42, 43, and 44. Representative commands are shown below; each command is repeated for the other seeds with the seed and checkpoint directory changed.

\begin{verbatim}
uv run python -m kanprey.train --model mlp --steps 8000 \
  --batch-size 32 --checkpoint-dir checkpoints/mlp_s42 --seed 42

uv run python -m kanprey.train --model kan --grid-size 2 --steps 8000 \
  --batch-size 32 --checkpoint-dir checkpoints/kan_grid2_s42 --seed 42

uv run python -m kanprey.train --model mlpedge --mlp-edge-hidden 8 \
  --steps 8000 --batch-size 32 \
  --checkpoint-dir checkpoints/mlpedge_h8_s42 --seed 42

uv run python -m kanprey.train --model grkan --steps 8000 \
  --batch-size 32 --checkpoint-dir checkpoints/grkan_corrected_s42 \
  --seed 42

uv run python -m kanprey.train --model basis --steps 8000 \
  --batch-size 32 --seed 42 --basis-family chebyshev \
  --basis-degree 3 --basis-groups 8 --basis-input-norm tanh \
  --checkpoint-dir checkpoints/basis_confirm/cheb_d3_g8_s42
\end{verbatim}

The unified seed-42 evaluation command used the same tokenizer, device, temperature 0.7, top-\(k\) 50, maximum 64 new tokens, and 16 prompts for every row. The prompt score is a heuristic pass/fail count and not a benchmark.

\subsection{Edge-audit sensitivity tables}

\begin{table}[htbp]
\centering
\caption{Seed repeat of aggregate B-spline KAN edge-audit metrics using 120 sample points per curve. The main text's 87.8\% nonlinear and 0.4\% inactive values come from the original 200-point audit; this repeat checks seed stability of the aggregate conclusion.}
\label{tab:edge-stability}
\begin{tabular}{@{}lcccc@{}}
\toprule
Seed & Median NLS & Nonlinear \(>0.1\) & Inactive \(\leq0.01\) & Min top-4 fPCA variance \\
\midrule
42 & 0.2927 & 88.34\% & 0.455\% & 99.924\% \\
43 & 0.2914 & 88.25\% & 0.455\% & 99.923\% \\
44 & 0.2929 & 88.43\% & 0.453\% & 99.923\% \\
\bottomrule
\end{tabular}
\end{table}

\begin{table}[htbp]
\centering
\caption{Nonlinearity-threshold sensitivity averaged over seeds 42--44 in the 120-point repeat audit.}
\label{tab:threshold}
\begin{tabular}{@{}lcccc@{}}
\toprule
Threshold & 0.05 & 0.10 & 0.15 & 0.20 \\
\midrule
Fraction of edges above threshold & 97.34\% & 88.34\% & 77.33\% & 66.80\% \\
Sample standard deviation & 0.014\% & 0.090\% & 0.159\% & 0.153\% \\
\bottomrule
\end{tabular}
\end{table}

\subsection{Cross-corpus edge audit and MLP pruning baseline}

\begin{table}[htbp]
\centering
\caption{Cross-corpus edge audit (Review~\#7, M3). Aggregate B-spline KAN grid-2 edge-function metrics on BabyLM Strict-Small (five seeds, 200-point reconstruction, identical code and settings to the GuppyLM audit), with a shared random-initialized control and the GuppyLM trained reference. Pooled over the six FFN layers and 884,736 edge functions per model. The trained BabyLM statistics match GuppyLM closely despite the different vocabulary (8,192 vs.\ 2,393), corpus, and objective; the random-init control confirms that closed-form approximability is training-induced while near-total top-four fPCA variance is basis-imposed at grid 2 on both corpora.}
\label{tab:babylm-audit}
\footnotesize
\setlength{\tabcolsep}{3pt}
\begin{tabular}{@{}lccccc@{}}
\toprule
Model (grid 2) & \shortstack{Median\\NLS} & \shortstack{Nonlin.\\\(>0.1\)} & \shortstack{Inactive\\\(\leq0.01\)} & \shortstack{Top-4\\fPCA} & \shortstack{Closed-form\\\(R^2{\geq}0.99\)} \\
\midrule
BabyLM trained (mean of 5 seeds) & 0.294 & 88.4\% & 0.44\% & 99.93\% & 92.8\% \\
BabyLM random init (control)     & 0.272 & 88.7\% & 0.52\% & 99.93\% & 64.3\% \\
GuppyLM trained (reference)      & 0.281 & 87.8\% & 0.40\% & 99.9\%  & 93.3\% \\
\bottomrule
\end{tabular}
\end{table}

\begin{table}[htbp]
\centering
\caption{MLP pruning baseline (Review~\#7, M4). Validation loss on the GuppyLM test split after zeroing the lowest-saliency fraction of the \(4\times\) GELU FFN hidden neurons (a structured prune that removes the matching fraction of FFN parameters), ranked by weight-magnitude saliency, data-driven activation magnitude, or at random. The KAN activity/random columns (grid-2 edges) are repeated from the pruning curve for comparison; note the unpruned baselines differ (MLP 0.270, KAN 0.282), so the fair comparison is the increase over each model's own baseline (Figure~\ref{fig:prune-compare}). A good MLP baseline (activation) tolerates comparable sparsity; KAN activity pruning is modestly gentler in \(\Delta\)-loss and far gentler than MLP weight-magnitude pruning, which collapses past 20\%.}
\label{tab:prune-mlp}
\begin{tabular}{@{}lccccc@{}}
\toprule
Fraction & MLP magnitude & MLP activation & MLP random & KAN activity & KAN random \\
\midrule
0\%  & 0.270 & 0.270 & 0.270 & 0.282 & 0.282 \\
10\% & 0.271 & 0.273 & 0.272 & 0.282 & 0.294 \\
20\% & 0.276 & 0.276 & 0.274 & 0.285 & 0.335 \\
30\% & 0.342 & 0.283 & 0.281 & 0.290 & 0.480 \\
40\% & 0.808 & 0.299 & 0.313 & 0.302 & 0.873 \\
50\% & 2.253 & 0.358 & 0.506 & 0.355 & 1.873 \\
\bottomrule
\end{tabular}
\end{table}

\subsection{Checkpoint lineage}

\begin{table}[htbp]
\centering
\caption{Checkpoint lineage for result rows.}
\label{tab:checkpoints}
\resizebox{\linewidth}{!}{%
\begin{tabular}{@{}l>{\raggedright\arraybackslash}p{0.66\linewidth}@{}}
\toprule
Result row & Checkpoints / artifacts \\
\midrule
GuppyLM MLP & \texttt{checkpoints/mlp\_s42}, \texttt{mlp\_s43}, \texttt{mlp\_s44} \\
GuppyLM KAN grid 2 & \texttt{checkpoints/kan\_grid2\_s42}, \texttt{kan\_grid2\_s43}, \texttt{kan\_grid2\_s44} \\
GuppyLM KAT grid 2 + attention & \texttt{checkpoints/kat\_s42}, \texttt{kat\_s43}, \texttt{kat\_s44} \\
GuppyLM \mlpedge{} \(h=5\) & \texttt{checkpoints/mlpedge\_s42}, \texttt{mlpedge\_s43}, \texttt{mlpedge\_s44} \\
GuppyLM \mlpedge{} \(h=8\) & \texttt{checkpoints/mlpedge\_h8\_s42}, \texttt{mlpedge\_h8\_s43}, \texttt{mlpedge\_h8\_s44} \\
GuppyLM corrected \grkan{} & \texttt{checkpoints/grkan\_corrected\_s42}, \texttt{grkan\_corrected\_s43}, \texttt{grkan\_corrected\_s44} \\
GuppyLM square \grkan{} & \texttt{checkpoints/basis\_confirm/grkan\_square\_p5q4\_g8\_s42} and seeds 43/44 \\
GuppyLM Chebyshev & \texttt{checkpoints/basis\_confirm/cheb\_d3\_g8\_s42} and seeds 43/44 \\
BabyLM KAN grid 2 (edge audit, Table~\ref{tab:babylm-audit}) & \texttt{checkpoints/babylm/kan\_grid2\_s42}--\texttt{s46} (five seeds) \\
Wikitext-103 MLP/\mlpedge{} & \texttt{checkpoints/scale/mlp/best.pt}, \texttt{checkpoints/scale/mlpedge\_h8/best.pt} \\
ClimbMix \grkan{} \(g=4\) & run \texttt{b-g4-20260610T193454Z-full-d12-grkan-g4}, checkpoint \texttt{model\_002520.pt} \\
ClimbMix stabilized \grkan{} \(g=4\) continuation & run \texttt{b-g4-stab-lr05-clip1-}\newline\texttt{cont3000to5040-sched5040-20260616T004144Z}, checkpoint \texttt{model\_005040.pt}; resumed from stabilized pilot step 3,000 with a reset 5,040-step learning-rate schedule \\
ClimbMix \grkan{} \(g=8\) & run \texttt{20260608T193716Z-d12-grkan-g8-corrected-r2}, checkpoint \texttt{model\_002520.pt} \\
ClimbMix MLP & \texttt{checkpoints/nanochat/d12-mlp/model\_002520.pt} plus matching tokenizer and eval artifacts \\
\bottomrule
\end{tabular}%
}
\end{table}

\section{Compute and runtime analysis}
\label{app:compute}

Parameter matching is necessary for architecture comparison, but it is not sufficient because different bases and topologies vary in arithmetic intensity, memory bandwidth, and kernel efficiency. We report validation loss, wall-clock, generation latency, and throughput separately rather than combining quality and time into a single score. The current experiments are not compute-matched, and a scalar such as loss divided by training time has no coherent efficiency interpretation when lower loss and lower runtime are both desirable.

Estimated forward-pass FLOPs per token for the GuppyLM FFN layers are:
\begin{itemize}
\item MLP-4x-GELU: two dense projections \(d_{\mathrm{model}}\!\to\!d_{\mathrm{ffn}}\!\to\!d_{\mathrm{model}}\), so \(\approx 4\,d_{\mathrm{model}}\,d_{\mathrm{ffn}} = 4\cdot384\cdot1536 = 2.36\times10^6\) FLOPs/token (the factor \(4\) is two projections times two FLOPs per multiply--add).
\item B-spline KAN grid 2: a single \kanlinear{} \(d_{\mathrm{model}}\!\to\!d_{\mathrm{model}}\) with \emph{no} \(4\times\) expansion, \(\approx 2\,d_{\mathrm{model}}^2\,(k+1) = 2\cdot384^2\cdot6 \approx 1.8\times10^6\) FLOPs/token (\(k=5\) spline terms plus the SiLU base). This is genuinely lower than the MLP because it omits the \(4\times\) expansion.
\item Chebyshev degree-3, \(g=8\) (\texttt{BasisKANFFN}): shares the MLP's two dense \(d_{\mathrm{model}}\!\to\!d_{\mathrm{ffn}}\!\to\!d_{\mathrm{model}}\) projections (\(2.36\times10^6\)) and adds only the grouped degree-3 recurrence over the \(d_{\mathrm{model}}\)- and \(d_{\mathrm{ffn}}\)-dimensional activations (\(\approx 10^4\)), totaling \(\approx 2.37\times10^6\) FLOPs/token---within \(\sim\)0.5\% of the MLP, not lower.
\item \mlpedge{} \(h=8\): \(\approx 2\,d_{\mathrm{model}}^2\,h = 2\cdot384^2\cdot8 = 2.36\times10^6\) FLOPs/token.
\item Rational \grkan{}, \(g=8\) (\texttt{GRKANFFN}): like the Chebyshev FFN it shares the MLP's two dense projections (\(2.36\times10^6\)) and adds the per-group Pad\'{e} rational recurrence (\(\approx 2\times10^4\)), totaling \(\approx 2.38\times10^6\) FLOPs/token.
\end{itemize}
These estimates are theoretical forward-pass FLOPs only; they omit backward-pass costs, activation memory traffic, and kernel-launch overhead. Crucially, the Chebyshev and rational \grkan{} FFNs share the MLP's two dense \(d_{\mathrm{model}}\!\to\!d_{\mathrm{ffn}}\!\to\!d_{\mathrm{model}}\) projections and differ only in the activation, so their theoretical FLOPs are within \(\sim\)1\% of the MLP rather than lower; only the B-spline KAN, which omits the \(4\times\) expansion, is genuinely cheaper. The slowdown is therefore not explained by arithmetic at all: at essentially matched FLOPs these variants run several times slower in the current Python/MPS implementation because the grouped/spline basis evaluation is not fused into optimized kernels. A \texttt{torch.profiler} breakdown (Table~\ref{tab:profiler}) quantifies this: the MLP achieves \(71.8\%\) model MFU on Apple M4 Pro/MPS while the Chebyshev model achieves \(14.1\%\).

\section{Stabilized GR-KAN continuation (unmatched)}
\label{app:continuation}

The admissible architecture comparison at 286M parameters is the matched 2,520-step ClimbMix endpoint of Table~\ref{tab:scale}, where both corrected \grkan{} group counts (\(g=4\) and \(g=8\)) trail the SwiGLU MLP baseline on validation BPB and DCLM CORE. To probe whether part of the corrected \grkan{} gap is an optimization-stability artifact rather than a capacity ceiling, we ran a deliberately stabilized \(g=4\) continuation. It resumes from a stabilized 3,000-step pilot, uses lower embedding, unembedding, and matrix learning rates, applies gradient clipping at 1.0, and resets the learning-rate schedule to a 5,040-step horizon. It reaches validation \(\bpb=0.8616\) at step 5,040, narrowing the gap against the available MLP checkpoint from \(+0.0440\) BPB at the matched 2,520-step endpoint to \(+0.0152\) BPB relative to the MLP's 2,520-step value.

\begin{figure}[htbp]
\centering
\includegraphics[width=0.82\linewidth]{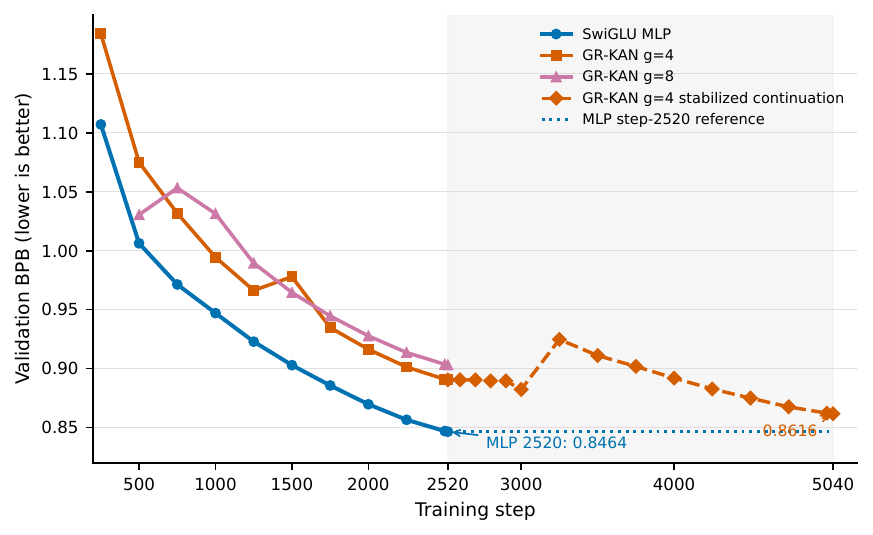}
\caption{Checkpoint-level validation BPB trajectories for the ClimbMix d12 stress test. The matched comparison runs through step 2,520, where both corrected \grkan{} group counts remain above the MLP trajectory. The shaded region shows the deliberately stabilized \(g=4\) continuation to step 5,040; the dotted line carries the MLP's 2,520-step endpoint forward only as a reference because no matched 5,040-step MLP continuation was run.}
\label{fig:scale-trajectories}
\end{figure}

This is \emph{not} an architecture comparison, and we do not treat it as one: the MLP baseline was not continued to 5,040 steps, the stabilization recipe and learning-rate schedule were changed, and the result is a single run. It is admissible only as an optimization-stability observation---evidence that the corrected \grkan{} keeps improving under gentler optimization---and it does not alter the matched-horizon conclusion. Figure~\ref{fig:scale-trajectories} separates the matched 2,520-step comparison from the shaded, unmatched stabilized extension and carries the MLP's 2,520-step endpoint forward only as a reference line. A credible scaling claim would still require matched longer-horizon baselines, repeated seeds, and ideally an intermediate-scale sweep to test whether the gap closes with more steps or more data.

\section{Additional interpretability figures}

\begin{figure}[htbp]
\centering
\includegraphics[width=\linewidth]{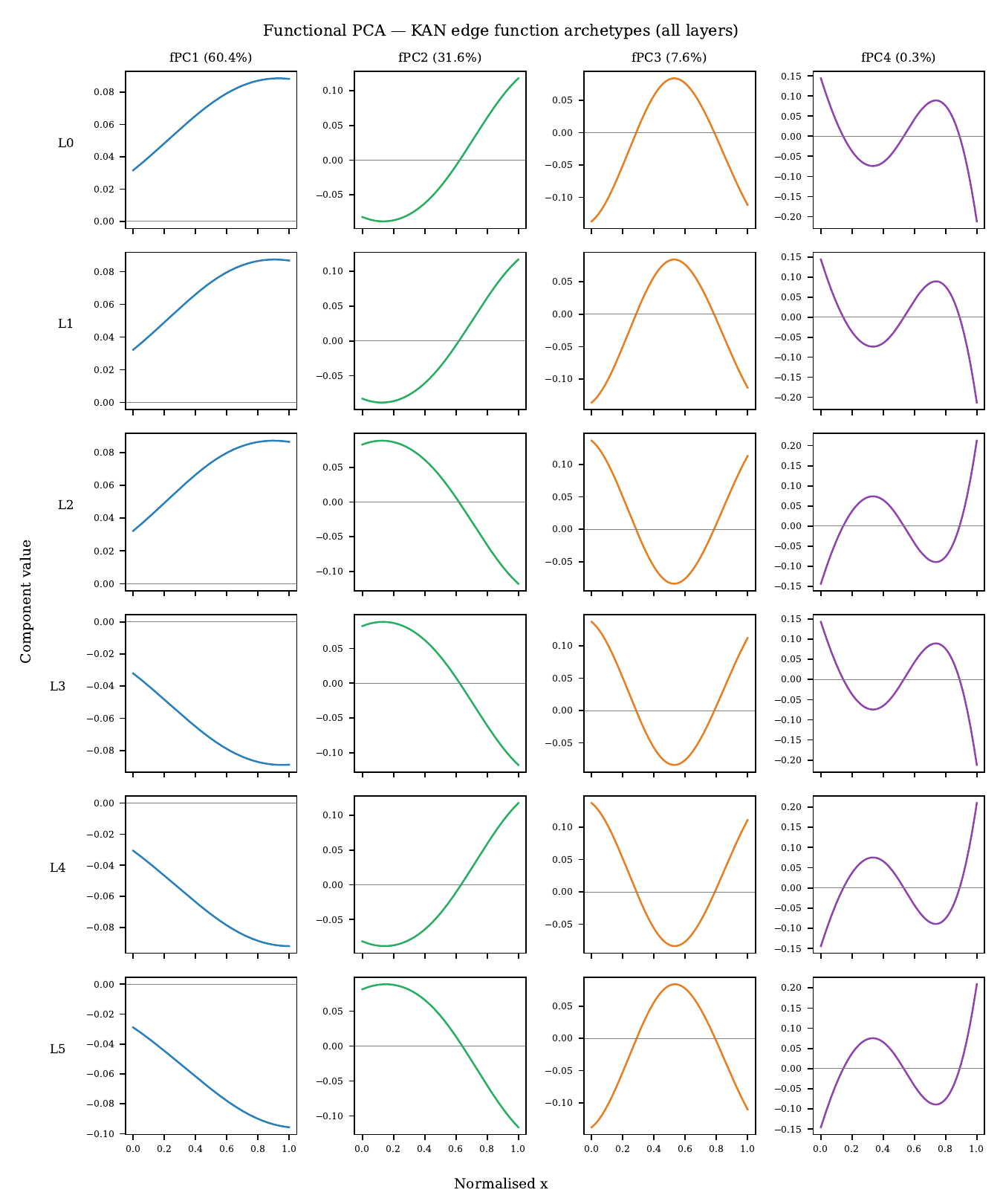}
\caption{Top four functional principal components across all six B-spline KAN FFN layers. Components explain 99.9\% of curve variance per layer in the original 200-point audit.}
\label{fig:fpca-all}
\end{figure}

\begin{figure}[htbp]
\centering
\includegraphics[width=\linewidth]{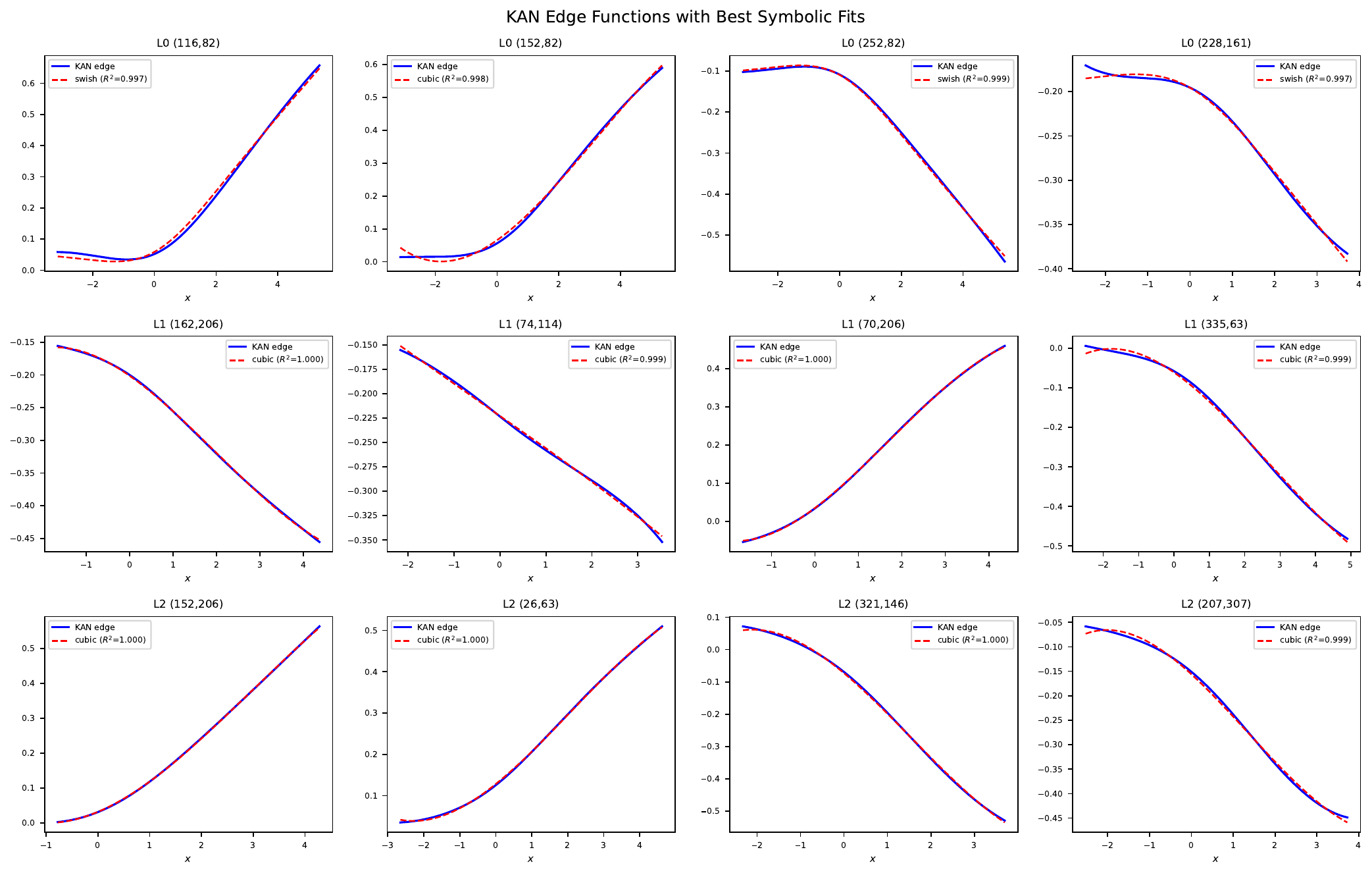}
\caption{Representative high-activity KAN edge functions and their best fits from the six-function smooth library. The fits illustrate closed-form approximability on the observed domain, not symbolic-law discovery.}
\label{fig:symbolic}
\end{figure}

\section{Additional statistical notes}
\label{app:stats}

We use ``benchmark-equivalent'' descriptively throughout: it means that no architecture shows a consistent advantage larger than the observed sub-1-point spread across BLiMP, the BLiMP supplement, and EWoK, with overlapping ten-seed confidence intervals and the only robust main-BLiMP effect reversing on the supplement. We do \emph{not} claim formal statistical equivalence under a pre-specified TOST or ROPE margin; a formal equivalence test would require declaring a region of practical equivalence (for example \(\pm1.0\) BLiMP point) and testing whether all critical pairwise differences fall inside it, which we leave to future work.

The BabyLM validation cross-entropy differences among the four critical architectures are, by contrast, all highly reliable across the ten seeds: every pairwise Welch test is significant at \(p<10^{-10}\), with the SwiGLU-, Chebyshev-, and \grkan{}-versus-MLP separations falling in the \(10^{-13}\) to \(10^{-20}\) range. This sharp contrast between the overwhelmingly significant validation-loss gaps and the sub-1-point, mostly non-significant BLiMP gaps is the statistical form of the paper's central dissociation. Because the same ten seeds (42--51) are reused across the critical architectures, a paired-by-seed analysis may be more appropriate than Welch independent-sample tests for some BLiMP comparisons; we report Welch tests as the primary analysis and flag paired-seed testing as a sensible robustness check, noting that the between-architecture BLiMP spread (sub-1-point) is comparable to the per-seed standard deviation (\(\approx\)0.6 points) regardless of pairing.

GuppyLM validation losses are reported as mean \(\pm\) sample standard deviation over three seeds. BabyLM benchmark accuracies are reported as mean \(\pm\) 95\% confidence interval over ten seeds for the four critical architectures (five for supporting rows, three for low-priority rows), with Welch two-sample \(t\)-tests against the MLP baseline; across the three non-MLP critical rows on the main BLiMP suite, only the canonical \grkan{} effect (\(p=0.005\)) survives a Bonferroni correction, and that effect reverses sign on the BLiMP supplement, so it does not support a directional claim. Wikitext-103 and ClimbMix stress tests are single-run comparisons; their gaps should be interpreted as case-study evidence until repeated. DCLM CORE per-task advantages are not used for standalone claims because per-task uncertainty and seed variation were not predeclared. The 16-prompt GuppyLM heuristic is no longer used as evidence, having been superseded by the standardized BabyLM benchmarks.

Statistical comparisons in the main text use Welch's two-sample \(t\)-test, which does not assume equal variances, with two-sided \(p\)-values. Bootstrap confidence intervals use 10,000 resamples with replacement and the percentile method. The low seed count (\(n=3\)) means that \(t\)-test power is limited and confidence intervals are wide; the tests are reported for completeness and to guide future replication with more seeds, not as definitive evidence of null or nonzero effects. In particular, with \(n=3\) the percentile bootstrap has poor coverage, so the reported intervals should be read as order-of-magnitude indications rather than calibrated 95\% intervals. We also do not apply a multiple-comparison correction across the several MLP-versus-variant tests; the comparisons are exploratory, and only the B-spline gap would survive any reasonable correction.

\section{Submission checklist and broader impact}

For a submission-facing version we summarize the standard checklist items in one place.

\begin{itemize}
\item \textbf{Code and data.} Code, experiment scripts, per-seed BabyLM evaluation reports, and small-scale checkpoints will be released in an anonymized archive before review and permanently archived on Zenodo with a citable DOI for the camera-ready version, including a provenance manifest (Data and artifact availability; Appendix~\ref{app:reprod}, Appendix~\ref{app:coderev}).
\item \textbf{Dataset licenses.} GuppyLM (\texttt{arman-bd/guppylm-60k-generic}) is MIT licensed; BabyLM Strict-Small is the public developmental corpus of the BabyLM challenge; Wikitext-103 is CC BY-SA 3.0/GFDL; ClimbMix traces to NVIDIA Nemotron-ClimbMix under CC BY-NC 4.0 for research use (Appendix Table~\ref{tab:data}).
\item \textbf{Compute.} Small-scale GuppyLM/BabyLM runs are on a single Apple M4 Pro (MPS); the Wikitext-103 scale run is on one RTX A6000; the 286M ClimbMix runs use an H100 for training and an RTX PRO 4500 for evaluation. Wall-clock, throughput, and hardware are reported per regime (Tables~\ref{tab:guppylm},~\ref{tab:profiler},~\ref{tab:scale}, Appendix~\ref{app:reprod}); peak VRAM and per-architecture FLOPs/token were not consistently logged and are listed as missing rather than inferred.
\item \textbf{Limitations.} The standardized evidence is well powered only at \(\approx\)13.8M-parameter scale; the larger \mlpedge{} and \grkan{} stress tests are single runs; EWoK is at chance at this model size; GLUE could not be run on the unmodified official pipeline; and the positive interpretability summaries are scoped to the low-capacity grid-2 basis (Section~\ref{sec:gridsweep}, Discussion).
\item \textbf{Human subjects / ethics.} The study involves no human subjects, no personally identifying data, and no biological or dual-use risk; all datasets are public.
\item \textbf{Broader impact.} Architecture papers can encourage adoption of inefficient variants when latency is underreported. This paper mitigates that risk by reporting negative latency and throughput evidence (Tables~\ref{tab:guppylm},~\ref{tab:profiler}) alongside the quality results, and by stating that the tested KAN-family FFNs are not demonstrated replacements for strong MLP baselines.
\end{itemize}

\end{document}